\documentclass{article}                     
\usepackage{graphicx}
\usepackage{amsmath}
\usepackage{amsfonts}
\usepackage{amssymb}
\usepackage{makeidx}
\usepackage{wrapfig}
\usepackage{hyperref}

\begin{document}

\title{Toward Automated Discovery of \\ Artistic Influence\footnote{The final publication is available at \href{http://dx.doi.org/DOI: 10.1007/s11042-014-2193-x}{Springer\textsuperscript{\textregistered}.}}}



\author{Babak Saleh \qquad
        Kanako Abe  \qquad
        Ravneet Singh Arora \\
        Ahmed Elgammal\\  \\
        Department of Computer Science
\\
Rutgers, The State University of New Jersey \\
\href{mailto:babaks@cs.rutgers.edu}{\texttt{\small{babaks,kanakoabe,rsingh,elgammal@cs.rutgers.edu}}}
}

%
%
%
%
\date{}
\maketitle

\begin{abstract}
Considering the huge amount of art pieces that exist, there is valuable information to be discovered. Examining a painting, an expert can determine its style, genre, and the time period that the painting belongs. One important task for art historians is to find influences and connections between artists. Is influence a task that a computer can measure? 
The contribution of this paper is in exploring the problem of computer-automated suggestion of influences between artists, a problem that was not addressed before in a general setting. We first present a comparative study of different classification methodologies for the task of fine-art style classification. A two-level comparative study is performed for this classification problem. The first level reviews the performance of discriminative vs. generative models, while the second level touches the features aspect of the paintings and compares semantic-level features vs. low-level and intermediate-level features present in the painting.  Then, we investigate the question ``Who influenced this artist?'' by looking at his masterpieces and comparing them to others. We pose this interesting question as a knowledge discovery problem. For this purpose, we investigated several painting-similarity and artist-similarity measures. As a result, we provide a visualization of artists (Map of Artists) based on the similarity between their works


\end{abstract}

\newpage

\section{Introduction}
\label{Sec:Intro}
\noindent
How do artists describe their paintings? They talk about their works using several different concepts. The elements of art are the basic ways in which artists talk about their works. Some of the {\em elements of art} include space, texture, form, shape, color, tone and line~\cite{lois}. Each work of art can, in the most general sense, be described using these seven concepts. Another important descriptive set is the {\em principles of art}. These include movement, unity, harmony, variety, balance, contrast, proportion, and pattern ~\cite{lois}. Other topics may include subject matter, brushstrokes, meaning, and historical context. As seen, there are many descriptive attributes in which works of art can be talked about.

\begin{figure}[t]
\centering
\includegraphics[width=\textwidth]{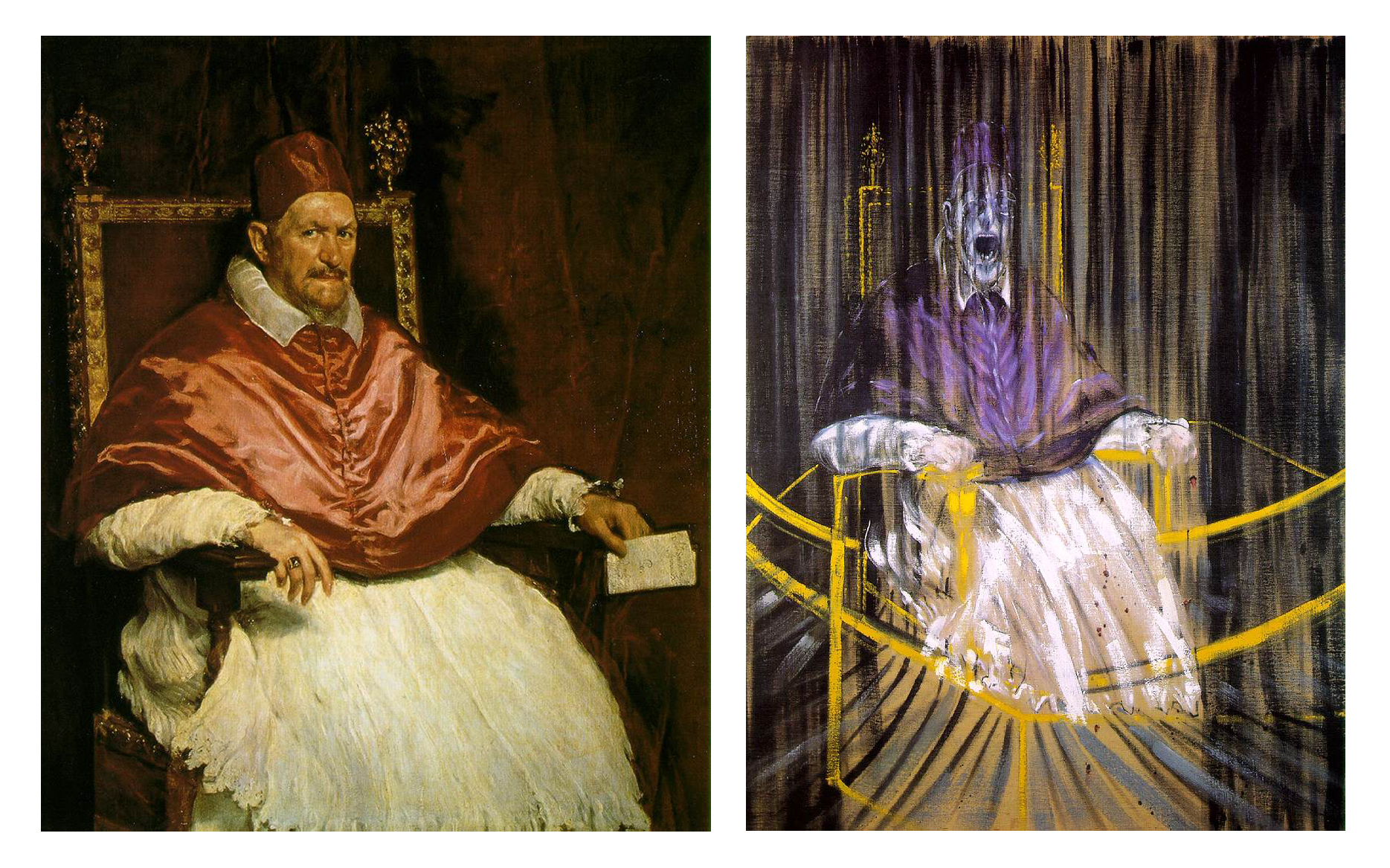}
\caption{An example of an often cited comparison in the context of influence. Left: Diego Vel\'azquez's Portrait of Pope Innocent X  (1650), and, Right: Francis Bacon's Study After Vel\'azquez's Portrait of Pope Innocent X (1953). Similar composition, pose, and subject matter but a different view of the work. 
}
\label{fig:figure1}
\end{figure}

One important task for art historians is to find influences and connections between artists. By doing so, the conversation of art continues and new intuitions about art can be made. An artist might be inspired by one painting, a body of work, or even an entire style of art. Which paintings influence each other? Which artists influence each other? Art historians are able to find which artists influence each other by examining the same descriptive attributes of art which were mentioned above. Similarities are noted and inferences are suggested. 

It must be mentioned that determining influence is always a subjective decision. We will not know if an artist was ever truly inspired by a work unless he or she has said so. However, for the sake of finding connections and progressing through movements of art, a general consensus is agreed upon if the argument is convincing enough. For example, Figure~\ref{fig:figure1} illustrates a commonly cited comparison for studying influence, in the work of Francis Bacon's Study After Vel\'azquez's Portrait of Pope Innocent X (1953), where similarity is clear in composition, pose, and subject matter.

Is influence a task that a computer can measure? In the last decade there have been impressive advances in developing computer vision algorithms for different object recognition-related problems including: instance recognition, categorization, scene recognition, pose estimation, etc. When we look into an image we not only recognize object categories, and scene category, we can also infer various aesthetic, cultural and historical aspects. For example, when we look at a fine-art paining, an expert, or even an average person can infer information about the style of that paining (e.g. Baroque vs. Impressionism), the genre of the painting (e.g. a portrait or a landscape), or even can guess the artist who painted it.  People can look at two painting and find similarities between them in different aspects (composition, color, texture, subject matter, etc.) This is an impressive ability of human perception for learning and judging complex aesthetic-related visual concepts, which for long have been thought not to be a logical process. In contrast, we tackle this problem using a computational methodology approach, to show that machines can in fact learn such aesthetic concepts.

Although there has been some research on automated classification of paintings e.g.~\cite{Arora12,CabralCDBC11,Carneiro11,Jia12,Graham10},  however, there is almost no  research done on computer-based measuring and determining of influence between artists. Measuring influence is a very difficult task because of the broad criteria for what influence between artists can mean. As mentioned earlier, there are many different ways in which paintings can be described. Some of these descriptions can be translated to a computer. For example, Li et al~\cite{Jia12} proposed automated way for analyzing brushstrokes to distinguish between Van Gogh and his contemporaries.  
For the purpose of this paper, we do not focus on a specific element of art or principle of art but instead we focus on finding and suggesting new comparisons by experimenting with different similarity measures and features.

What is the benefit of the study of automated methods of analyzing painting similarity and artistic influences? By including a computer quantified judgement about which artists and paintings may have similarities, it not only finds new knowledge about which paintings are connected using a mathematical criteria, but also keeps the conversation going for artists. It challenges people to consider possible connections in the timeline of art history that may have never been seen before. {\em We are not asserting truths but instead suggesting a possible path towards a difficult task of measuring influence.}

Besides the scientific merit of the problem, there are various application-oriented motivations. With the increasing volumes of digitized art databases on the internet comes the daunting task of organization and retrieval of paintings. There are millions of paintings present on the internet. To manage properly the databases of these paintings, it becomes very essential to classify paintings into different categories and sub-categories. This classification structure can be utilized as an index and thus can improve the speed of retrieval process. Also it will be of great significance if we can infer new information about an unknown painting using already existing databases of paintings, and as a broader view can infer high-level information like influences between painters.

\begin{figure}[t]
\centering
\includegraphics[width=1\linewidth]{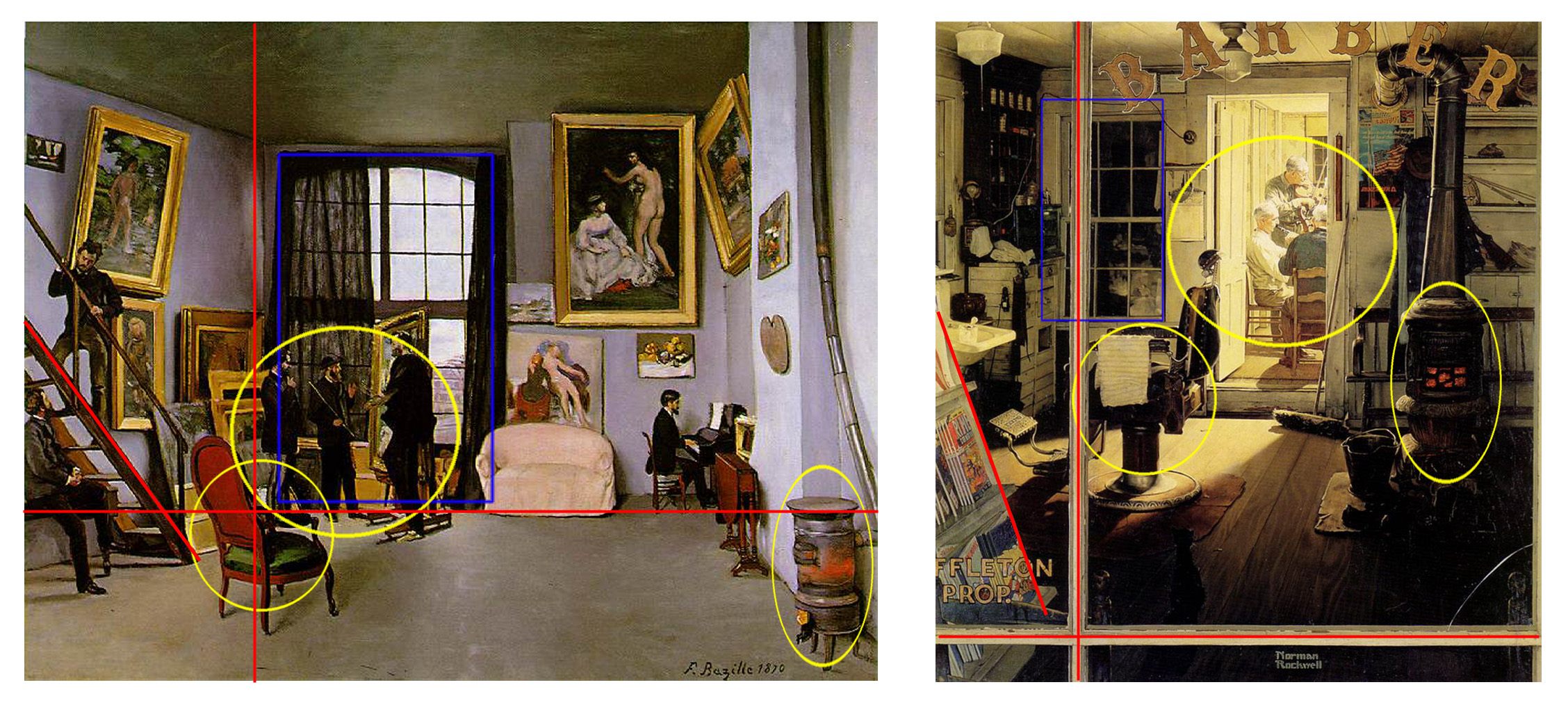}
\caption{Fr\'ed\'eric Bazille's \textit{Studio 9 Rue de la Condamine} (left) and Norman Rockwell's \textit{Shuffleton's Barber Shop} (right). The composition of both paintings is divided in a similar way. Yellow circles indicate similar objects, red lines indicate composition, and the blue square represents similar structural element. The objects seen -- a fire stove, three men clustered, chairs, and window are seen in both paintings along with a similar position in the paintings. After browsing through many publications and websites, we conclude that this comparison has not been made by an art historian before.}
\label{fig:figure2}
\end{figure}

Although the meaning of a painting is unique to each artist and is completely subjective, it can somewhat be measured by the symbols and objects in the painting. Symbols are visual words that often express something about the meaning of a work as well. For example, the works of Renaissance artists such as Giovanni Bellini and Jan Van-Eyck use religious symbols such as a cross, wings, and animals to tell stories in the Bible. This shows the need for an object-based representation of images. We should be able to describe the painting from a list of many different object classes. By having an object-based representation, the image is described in a high-level semantic as opposed to low-level features such as color and texture, which facilitates suggesting influences based on subject matter.  Paintings do not necessarily have to look alike, but if they do, or have reoccurring objects (high-level semantics), then they might be considered similar. If influence is found by looking at similar characteristics of paintings, the importance of finding a good similarity measure becomes prominent. Time is also an essential factor in determining influence. An artist cannot influence another artist in the past. Therefore the linearity of paintings cuts down the possibilities of influence.

The contribution of this paper is in exploring the problem of computer-automated suggestion of influences between artists, a problem that was not addressed before in a general setting. From a machine-learning point of view, we approach the problem as an unsupervised knowledge discovery problem.  
Our methodology is based on three components: 1) studying different representations of painting to determine which is more useful for the task of influence detection; 2) measuring similarity between paintings; 3) studying different measures of  similarity between artists.
We collected a comprehensive painting dataset for conducting our study. The data set contains 1710 high-resolution images of paintings by 66 artist spanning the time period of 1412-1996 and containing 13 painting styles.
We also collect a ground-truth data set for the task of artistic influences, which mainly contains positive influences claimed by art historian. This ground-truth is only used for the overall evaluation of our discovered/suggested influences, and is not used in the learning or knowledge-discovery.

We hypothesis that a high-level semantic representation of painting would be more useful for the task of influence detection. 
However, evaluating such a hypothesis requires comparing the performance of different features and representation in detecting influences against a ground-truth of artistic influences, containing both positive and negative example.  However, because of the limited size of the available ground-truth data, and the lack of negative examples in it, it is not useful for comparing different features and representations. Instead we resort to a highly correlated task, which is classifying painting style. The hypothesis is that features and representations that are good for style classification (which is a supervised learning problem), would also be good for determining influences (which is an unsupervised problem).  Therefore, we performed a comprehensive comparative study of different features and classification models for the task of classifying painting style among seven different styles. This study is described in details in Sec~\ref{Sec:classification}. The conclusion of this study confirms our hypothesis that high-level semantic features would be more useful for the task of style classification, and hence useful for determining influences.


Using the right features to represent the painting paves the way to judge similarity between paintings in a quantifiable way.
Figure~\ref{fig:figure2} illustrates an example of similar paintings detected by our automated methodology;  Fr\'ed\'eric Bazille's \textit{Studio 9 Rue de la Condamine} (1870) and Norman Rockwell's \textit{Shuffleton's Barber Shop} (1950).   After browsing through many publications and websites, we concluded, to the best of our knowledge, that this comparison has not been made by an art historian before. The painting might not look similar at the first glance, however, a closer look reveals striking similarity in composition and subject matter, that is detected by our automated methodology (see caption for details). Other example similarity can be seen in Figures~\ref{fig:go} \&~\ref{fig:braque}. 

Measuring similarity between painting is fundamental to discover influences, however, it is not clear how painting similarity might be used to suggest influences between artist. The paintings of a given artist can span extended period of time and can be influenced by several other contemporary and prior artists. Therefore, we investigated several artist distance measures to judge similarity in their work and suggest influences. As a result of this distance measures, we can achieve visualizations of how artists are similar to each other, which we denote by a map of artists.



The paper is structured as follows: Section~\ref{Sec:Rel} provides a literature survey on the topic of computer-based methods for analyzing painting. Section~\ref{Sec:Dataset} describes the data set used in our study. Section~\ref{Sec:classification} describes our comparative study for the task of painting style classification, including the methodologies, features and the results. Section~\ref{Sec:Prob_def} describes our methodology for judging artistic influence. Section~\ref{Sec:influence_disc} represents qualitative and quantitative evaluation of our automated influence study.



\section{Related Works}
\label{Sec:Rel}
There is little work done in the area of automated fine-art classification. Most of the work done in the problem of paintings classification utilizes low-level features such as color, shades, texture and edges. Lombardi~\cite{Lombardi} presented a comprehensive study of the performance of such features for paintings classification. In that paper the style of the painting was identified as a result of recognizing the painter. Sablatnig et al.~\cite{sablatnig} used brushstroke patterns to define structural signature to identify the artist style. Khan et al.~\cite{khan} used a Bag of Words (BoW) approach with low-level features of color and shades to identify the painter among eight different artists. In~\cite{robert} and~\cite{widjaja} similar experiments with low-level features were conducted. Unlike most of the previous works that focused on inferring the artist from the painting, our goal is to directly recognize the style of the painting, and discover artist similarity and influences, which are more challenging tasks. 

Carneiro et al.~\cite{Carneiro12} recently published the dataset ``PRINTART" on paintings along with primarily experiments on image retrieval and painting style classification. They provided three levels of annotation for the ``PRINTART" dataset: Global, Local and Pose annotation. However this dataset contains only monochromatic artistic images. We present a new dataset which has chromatic images and its size is about double the ``PRINTART" dataset covering a more diverse set of styles and topics. Carneiro et al.~\cite{Carneiro12} showed that the low-level texture and color features exploited for photographic image analysis are not as effective because of inconsistent color and texture patterns describing the visual classes (e.g. humans) in artistic images.  

Carneiro et al.~\cite{Carneiro12} define artistic image understanding as a process that receives an artistic
image and outputs a set of global, local and pose annotations. The global annotations consist of a set of artistic keywords describing the contents of the image. Local annotations comprise a set of bounding boxes that localize certain visual classes, and pose annotations consisting of a set of body parts that indicate the pose of humans and animals in the image. Another process involved in the artistic image understanding is the retrieval of images given a query containing an artistic keyword. In.~\cite{Carneiro12} an improved inverted label propagation method was proposed that produced the best results, both in the automatic (global, local and pose) annotation and retrieval problems.

Carneiro et. al.~\cite{Carneiro11} targeted the problem of annotating an unseen image with a set of global labels, learned on top of annotated paintings. Furthermore, for a given set of visual classes, they are able to retrieve the painting which shows the same characteristics. They have proposed a graph-based learning algorithm based on the assumption that visually similar paintings share same annotation. They formulated the global annotation problem with a combinatorial harmonic approach, which computes the probability that a random walk starting at the test image 
first reaches each of the database samples. However all the samples are from fifteen to seventeen century and focused on religious themes.

Graham et. al.~\cite{Graham10} posed the question of finding the way we perceive two artwork as similar to each other. Toward this goal, they acquired strong supervision of human experts to label similar paintings. They apply multidimensional scaling methods to paired similar paintings from either Landscape or portrait/still life and showed that similarity between paintings can be interpreted as basic image statistics. In the experiments they show that for landscape paintings, basic grey image statistics is the most important factor for two artwork to be similar. For the case of still life/portrait most important elements of similarity are semantic variables, for example representation of people. 

Unlike the case of ordinary images, where color and texture are proper low-level features to be used for a diverse set of tasks (e.g. classification), these might not describe paintings well. Color and texture features are highly prone to variations during digitization of paintings. In the case of color, it also lacks fidelity due to aging. The effect of digitization on the computational analysis of paintings is investigated in great depth by Polatkan et. al~\cite{brdahujapo09}.

The aforementioned reasons make the brushstrokes more meaningful features for describing paintings. Li et al.~\cite{Jia12} used fully automatic extracted brushstrokes to describe digitized paintings. Their novel feature extraction method is developed by the integration of edge detection and clustering-based segmentation. Using these features they found that regularly shaped brushstrokes are tightly arranged, creating a repetitive and patterned impression that can represent Van Gogh style and help to distinguish his work from his contemporaries. They have conducted a set of analysis based on 45 digitized oil paintings of Van Gogh from museum's collections. Due to small number of samples, and to avoid overfitting, they state this problem as a hypothesis testing rather than classification. They hypothesize which factors are eminent in Van Gogh style comparing to his contemporaries and tested them by statistical approaches on top of brushstroke features. 

Cabral et al~\cite{CabralCDBC11} approached the problem of ordering paintings and estimating their time period. They formulated this problem as embedding paintings into a one dimensional manifold and tried two different methods: on one hand, they applied unsupervised embedding using Laplacian Eignemaps~\cite{Belkin02laplacianeigenmaps}. To do so they only need visual features and defined a convex optimization to map paintings to a manifold. This approach is very fast and do not need human expertise, but its accuracy is low.  On the other hand, since some partial ordering on paintings is available by experts, they use these information as a constraint and used Maximum Variance Unfolding~\cite{weinberger2004learning} to find a proper space, capturing more accurate ordering of paintings.

\section{Dataset}
\label{Sec:Dataset}
Our dataset contains a total of 1710 images of art works by 66 artists, chosen from Mark Harden's Artchive database of fine-art~\cite{artchive}. Each image is annotated with the artist's first name, last name, title of work, year made, and style. The majority of the images are of the full work while a few are details of the work. We are primarily dealing with paintings but we have included very few images of sculptures as well. The artist with the largest number of images is Paul C\'ezanne with 140 images, and the artist with the least number of works is Hans Hoffmann with 1 image.

The artists themselves ranged from 13 different styles throughout art history. These include, with no specific order, Expressionism (10 artists), Impressionism (10), Renaissance (12), Romanticism (5), Cubism (4), Baroque (5), Pop (4), Abstract Contemporary (7), Surrealism (2), American Modernism (2), Post-Impressionism (3), Symbolism (1), and Neoclassical (1). The number in the parenthesis refers to the number of artists in each style category. Some styles were condensed such as \textit{Abstract Contemporary}, which includes works in the \textit{Abstract Expressionism, Contemporary,} and \textit{De Stijl} periods. The \textit{Renaissance} period has the most images (336 images) while \textit{American Modernism} has the least (23 images).  The average number of images per style is 132.
\begin{figure}[h]
\includegraphics[width=1\linewidth]{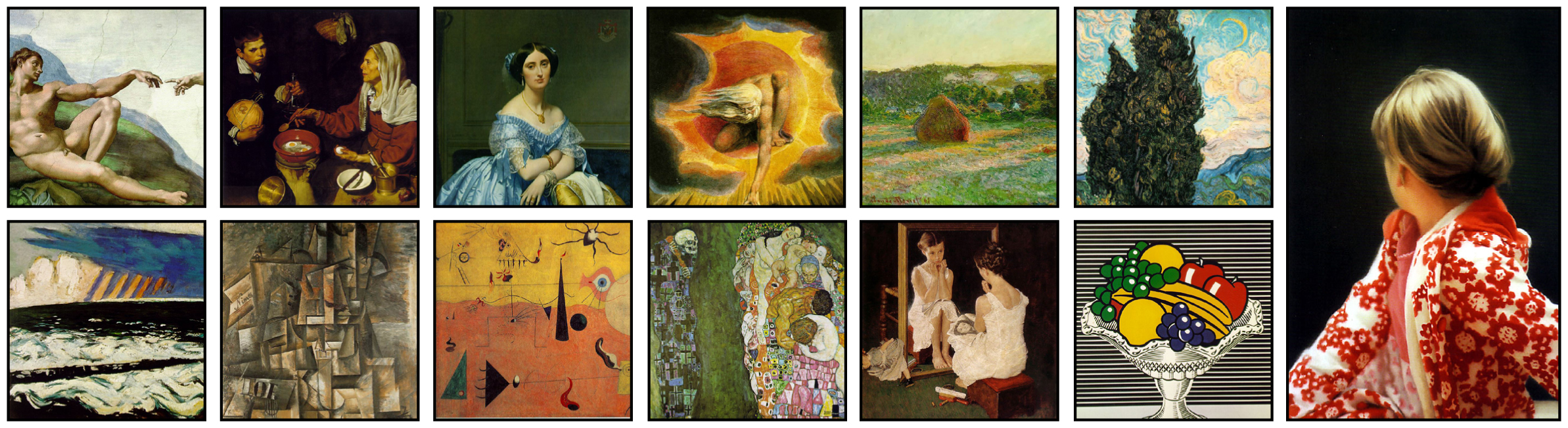}
\caption{Examples of paintings from thirteen styles: Renaissance, Baroque, Neoclassical, Romanticism, Impressionism, Post-Impressionism, Expressionism, Cubism, Surrealism, Symbolism, American Modernism, Pop, and Abstract Contemporary.}
\label{fig:examples}
\end{figure}
The earliest work is a piece by Donatello in 1412, while the most recent work is a self portrait by Gerhard Richter done in 1996. The earliest style is the \textit{Renaissance} period with artists like Titian and Michelangelo during the 14th to 17th century. As for the most recent style, art movements tend to overlap more in recent years. Richter's painting from 1996 is in the \textit{Abstract Contemporary} style. 

\section{Painting-Style Classification: A Comparative Study}
\label{Sec:classification}

In this section we present the details of our study on painting style classification.
The problem of painting style classification can be stated as: Given a set of paintings for each painting style, predict the style of an unknown painting. A lot of work has been done so far on the problem of image category recognition, however the problem of painting classification proves quite different than that of image category classification. Paintings are differentiated, not only by contents, but also by style applied by a particular painter or school of painting or by the age when they were painted. This makes painting classification problem much more challenging than the ordinary image category recognition problem.

In this study we will approach the problem of painting style classification from a supervised learning perspective. A two-level comparative study is conducted for this classification problem. The first level reviews the performance of discriminative vs. generative models, while the second level touches the feature aspects of the paintings and compares semantic-level features vs. low-level and intermediate-level features present in the painting. 

For experimental purposes seven fine-art styles are used, namely {\it Renaissance, Baroque, Impressionism, Cubism, Abstract, Expressionism, and Popart.} 
Various different sets of comparative experiments were performed focused on evaluation of classification accuracy for each methodology. We evaluated three different methodologies, namely: 
\begin{enumerate}
\item Discriminative model using a Bag-of-Words (BoW) approach
\item Generative model using BoW
\item Discriminative model using Semantic-level features
\end{enumerate}

\begin{figure}
\label{fig:chart}
\centering
\includegraphics[width=\linewidth]{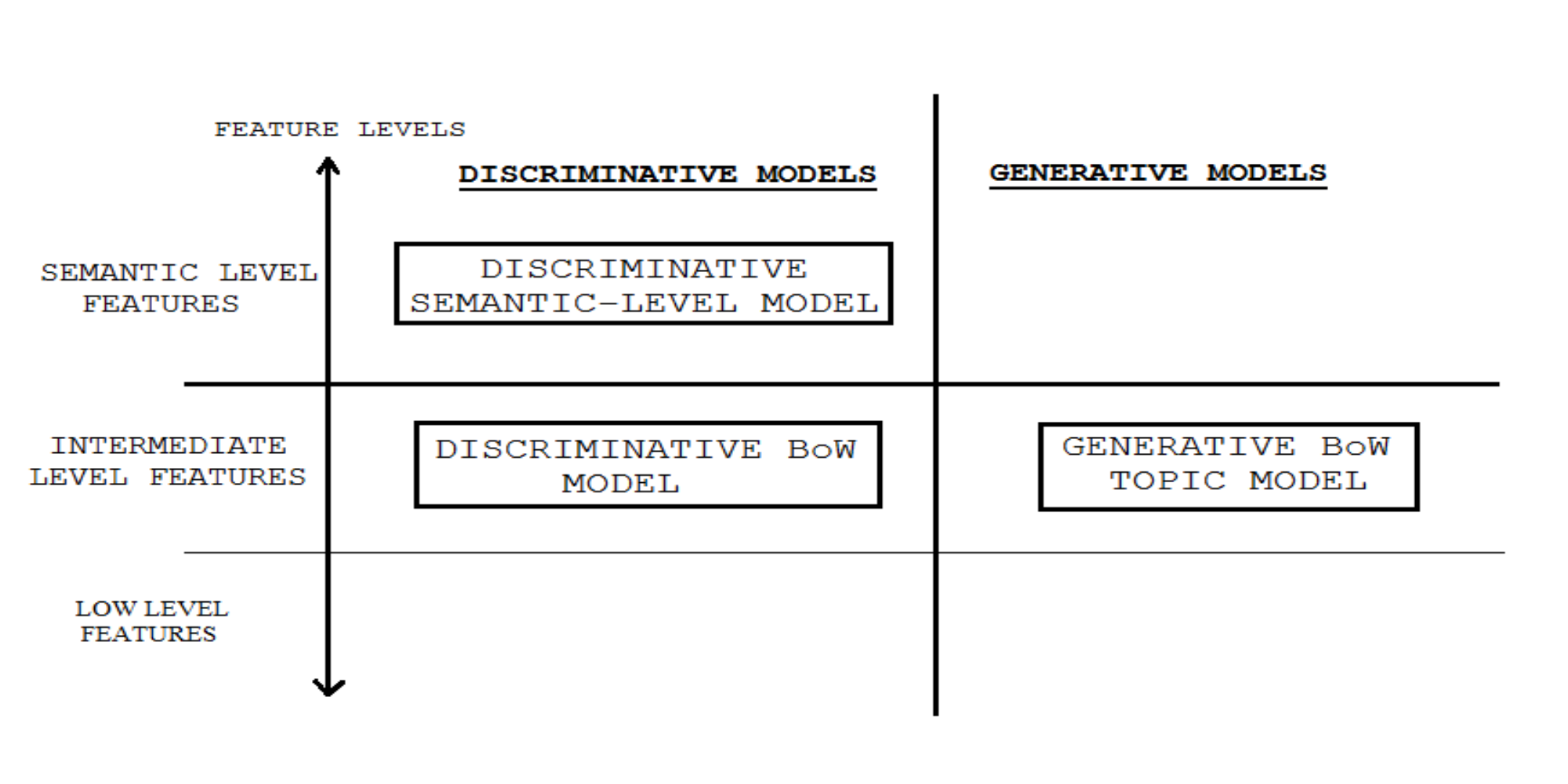}
\caption{Illustrative diagram of approaches for style classification of fine art paintings}
\end{figure}

As shown in Figure~\ref{fig:chart}, these three models differ in terms of the classification methodology, as well as the type of features used to represent the painting.  The Discriminative Semantic-Level model applies a discriminative machine learning model upon features capturing semantic information present in a painting, while Discriminative and Generative BoW models employs discriminative and generative machine learning models, respectively, on the Intermediate level features represented using a BoW model. 

A generative model has the property that it specifies a joint probability distribution over observed samples and their labels. In other words, a generative classifier learns a model of joint probability distribution $p(x,y)$, where $x$ denotes the observed samples and $y$ are the labels. Bayes rule can be applied to predict the label $y$ for a given new sample $x$, which is determined by the probability distribution $p(y|x)$. Since a generative model calculates the distribution $p(x|y)$ as an intermediate step, these can be used to generate random instances $x$ conditioned on target labels $y$.  A discriminative model, in contrast, tries to estimate the distribution $p(y|x)$ directly from the training data. Thus, a discriminative model bypasses the calculation of joint probability distribution $p(x,y)$ and avoids the use of Bayes rule. We refer the reader to~\cite{Ng2001} for a comprehensive comparison of both learning models.

It is also very important to make distinction between Low, Intermediate, and Semantic -level features at this stage.  Low-level features capture directly the formal characteristics of paintings such as color, texture, edges, light etc. The average intensity of all the pixels, color histogram representing color composition of paintings and number of edges are examples of  low-level features that capture the formal elements light, color and edges respectively.   Intermediate-level features apply local-level descriptors like SIFT~\cite{SIFT} and CSIFT~\cite{CSIFT} on various regions of an image. Local level descriptors instead of summarizing the whole image, represents localized regions of an image. A Bag of Words model is applied to generate an intermediate representation of the image. A Bag of Words model first creates fixed number of clusters from the localized regions of all the images (a codebook of visual vocabulary) and further represents each image by the histogram capturing the frequency of the code words in that image. 
Semantic-level features capture the semantic content classes such as water, sand, cars etc. present in an image. Thus, such frequency of semantic classes can help us in ranking images according to their semantic similarity. A feature vector where each element denotes the probability of existence of a semantic class is an example of semantic feature. 
It is worth noting that, instead of using low-level features like color, light, shades and texture our study is focused on intermediate-level features (BoW features) and semantic-level features.  

We hypothesize the following claims Ð 1) Semantic-level information contained in a painting can be very well utilized for the task of classification and 2) Generative models like Topic models are very much capable of capturing the thematic structure of a painting. It is easy to visualize a topic or theme in the case of documents. For documents, a topic can be a collection of particular set of words. For example, a science topic is characterized by the collection of words like atom, electrons, protons etc. For images represented by a Bag of Word model, each word is represented by the local level descriptor used to describe the image. Thus a collection of particular set of such similar regions can constitute a topic. For example, collection of regions representing mainly straight edges can constitute the topic trees. Similarly, set of regions having high concentration of blue color can form up a theme related to sky or water.

The following subsections describe the details of the compared methodologies.


\subsection{Discriminative Bag-of-Words model}
Bag of Words(BoW)~\cite{BoW} is a very popular model in text categorization to represent documents, where the order of the words does not matter. BoW was successfully adapted for object categorization, e.g. in~\cite{Csurka2004,BoW3D,BoWscene}. Typical application of BoW on an image involves several steps, which includes:
\begin{description} 
\item[]1) Locating interest points in an image 
\item[]2) Representation of such points/regions using feature descriptors 
\item[]3) Codebook formation using K-Means clustering, to obtain a ``dictionary'' or a codebook of visual words. 
\item[]4) Vector quantization of the feature descriptor; each descriptor is encoded by its nearest visual word from the codebook. 
\item[]5) Generate an intermediate-level representations for each image using the codebook, in the form of a histogram of the visual words present in each image. 
\item[]6)  Train a discriminative classifier on the intermediate training feature vectors for each class.
\item[]7) For classification, the trained classifier is applied on the BoW feature vector of a test image.
\end{description}

Thus, the end result of a Bag of Words model is a histogram of words, which is used as an intermediate-level feature to represent a painting. In our study, we applied a Support Vector Machine  (SVM) classifier~\cite{SVM} on a code-book trained on images from our dataset. We used two variant of the widely used Scale Invariant Feature Transform ``SIFT" features~\cite{SIFT} called Color SIFT (CSIFT)~\cite{CSIFT} and opponent SIFT (OSIFT)~\cite{osift} as local features. The SIFT~\cite{SIFT} is invariant to image scale, rotation, affine distortion and illumination. It uses edge orientations to define a local region and also utilizes the gradient of an image. Also, the SIFT descriptor is normalized and hence is also immune to gradient magnitude changes. CSIFT~\cite{CSIFT} and opponent SIFT (OSIFT)~\cite{osift} extends SIFT features for color images, which is essential for the task of painting-style classification.  In an earlier study by Van De Sande et al~\cite{VanDeSande2010} opponent SIFT was shown to outperform other color SIFT variants in image categorization tasks.

\subsection{Discriminative Semantic-level model}
In this approach a discriminative model is employed on top of semantic-level features. Seeking semantic-level features, we extracted the Classeme feature vector~\cite{aleb} as the visual feature for each painting. Classeme features are output of a set of classifiers corresponding to a set of $C$ category labels, which are drawn from an appropriate term list, defined in~\cite{aleb}, and not related to our fine-art context. For each category $c \in \lbrace 1 \cdots C \rbrace $, a set of training images was gathered by issuing a query on the category label to an image search engine. After a set of coarse feature descriptors (Pyramid HOG, GIST) is extracted, a subset of feature dimensions was selected~\cite{aleb}. Using this reduced dimension features, a one-versus-all classifier $\phi_{c}$ is trained for each category. The classifier output is real-valued, and is such that $\phi_{c}(x) > \phi_{c}(y)$ implies that $x$ is more similar to class $c$ than $y$ is. Given an image $x$,  the feature vector (descriptor) used to represent it is the Classeme vector $  [\phi_{1} (x),  \cdots, \phi_{C} (x)]$. The Classeme feature is of dimensionality $N = 2569$.

We used such feature vectors to train a Support Vector Machine (SVM)~\cite{SVM} classifier for each painting genre. We hypothesize that Classeme features are suitable for representing and summarizing the overall contents of a painting since it captures semantic-level information about object presence in a painting encoded implicitly in the output of the pre-trained classifiers.

\subsection{Generative Bag-of-Words Topic model}
Generative topic model uses Latent Dirichlet Allocation (LDA)~\cite{blei}. In studies~\cite{fei} and~\cite{sivic}, LDA and Probabilistic Latent Semantic Analysis (pLSA) topic models have been applied for object categorization, localization and scene categorization. This paper is the first evaluation of such models in the domain of fine-art categorization.

\begin{figure}
\label{fig:gm}
\centering
\includegraphics[scale=.45]{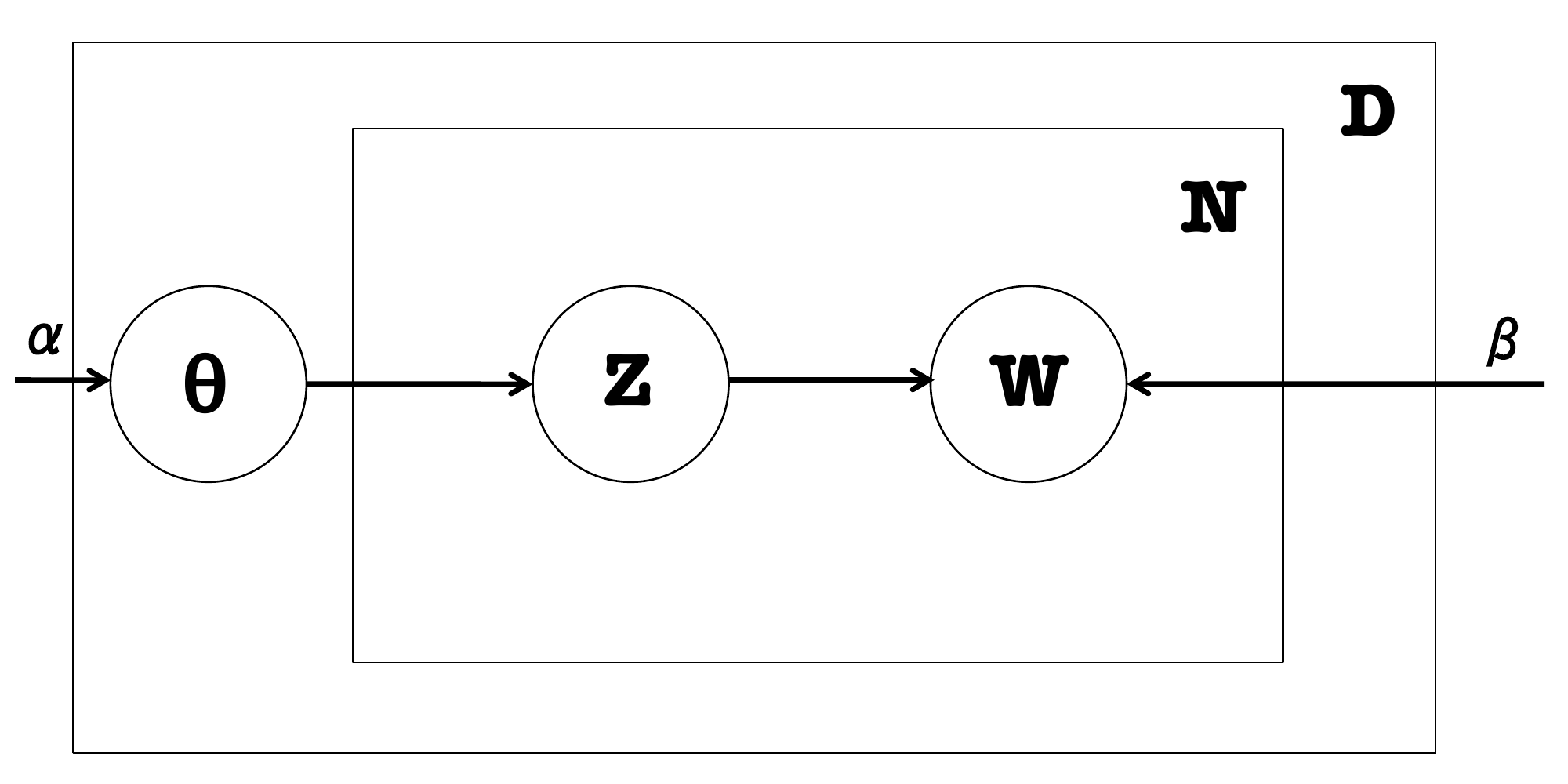}
\caption{Graphical model representing Latent Dirichlet Allocation}
\end{figure}

For the purpose of our study, we used Latent Dirichlet Allocation (LDA~\cite{blei}) topic model and applied it on BoW representation of paintings using both CSIFT and OSIFT features. In LDA, each item is represented by a finite mixture over a set of topics and each topic is characterized by a distribution over words. Figure~\ref{fig:gm} shows a graphical model for the image generation process. As shown in the model, parameter $\Theta$ defines the topic distribution for each image (total number of images is D.) $\Theta$ is determined by Dirichlet parameter $\alpha, \beta$  and represents the word distribution for each topic. The total number of words is N.  
 To use LDA for the classification task, we build model for each of the styles in our framework. First step is to represent each training image by a quantized vector using Bag-of-Words model described earlier. This vector quantized representation of each image is used for parameter estimation using Variational Inference. Thus, we will get LDA parameters $\Theta_c$ and $\beta_c$ for each category $c$. Once we have a new test image, $d$, we can infer the parameter $\Theta_{cd}$ for each category and $p(d| \Theta_{cd}, \beta_c)$ is used as the likelihood of the image belonging to a particular class $c$.
 
 \subsection{Style Classification Results}
For the task of Style classification of paintings, we focus on a subset of our dataset that contains seven categories of paintings namely Abstract, Baroque, Renaissance, Pop-art, Expressionism, Impressionism and Cubism. Each category consists of 70 paintings. For each of the following experiments five-fold cross-validation was performed, with 20$\%$ of the images chosen for testing purpose in each fold.

\begin{table}[t]
	\begin{center}
	\resizebox{\textwidth}{!}{
    \begin{tabular}{ | l | l | l | l | l | l | l | l |}
    \hline
     Confusion(\%) & Baroque &  Abstract & Renaissance & Pop-Art & Expressionism & Impressionism & Cubism  \\ \hline
     Baroque & 87.5 & 0 & 14.3 & 0 & 5.3 & 17.8 & 1.78 \\ \hline
     Abstract & 0 & 64 & 0 & 7.1 & 7.1 & 1.8 & 1.9 \\ \hline
	 Renaissance & 5.4 & 0 & 64.3 & 5.35 & 14.3 & 3.5 & 1.8 \\ \hline	
	 Pop-Art & 0 & 1.78 & 1.8 & 73.1 & 0 & 3.5 & 1.8 \\ \hline
	 Expressionism & 1.8 & 20.2 & 7.1 & 3.6 & 48.2 & 17.8 & 12.9 \\ \hline
	 Impressionism & 5.36 & 8 & 9 & 5.3 & 17.8 & 48.2 & 9.2 \\ \hline
	 Cubism & 0 & 6 & 3.5 & 5.3 & 7.1 & 7.1 & 72.4 \\ \hline
    \end{tabular}}
	\end{center}
    \caption{Confusion matrix for Discriminative Semantic Model}
    \label{tab:DisSemMod}
\end{table}

For codebook formation, Harris-Laplace detector~\cite{ShiHD09} is used to find the interest points. For efficient computation the number of interest points for each painting is restricted to 3000. Standard K-means Clustering algorithm is used to build a Codebook of size 600 words. SVM classifier is trained on both intermediate-level and semantic-level descriptors. For SVM, we use Radial Basis function (RBF) kernels. To determine parameters for the SVM, the grid search algorithm implemented by~\cite{libsvm} is employed. Grid search algorithm uses cross-validation to pick up the optimum parameter values. Also this process is preceded by scaling of dataset descriptors. For experiments with LDA, David Beli's C-code~\cite{blei} is used for the task of parameter estimation and inference. This C-code uses Variational Inference technique, which tries to estimate parameters $\beta$ and $\Theta$ using a similar and simpler model. For parameter estimation alpha is set to be $0.1$ and LDA code is set to estimate the value of $\alpha$ during the estimation process.

\begin{table}
	\begin{center}
	\resizebox{\textwidth}{!}{
    \begin{tabular}{ | l | l | l | l | l | l | l | l |}
    \hline
     Confusion(\%) & Baroque &  Abstract & Renaissance & Pop-Art & Expressionism & Impressionism & Cubism  \\ \hline
     Baroque & 71.4 & 0 & 12.9 & 0 & 8.5 & 17.1 & 0 \\ \hline
     Abstract & 0 & 48 & 5.8 & 10 & 8.5 & 5.7 & 7.1 \\ \hline
	 Renaissance & 18.6 & 6.7 & 41.4 & 0 & 5.8 & 9.3 & 18.5 \\ \hline	
	 Pop-Art & 0 & 15 & 0 & 70 & 11.5 & 9.3 & 15.7 \\ \hline
	 Expressionism & 0 & 15 & 18.6 & 2.8 & 28.5 & 12.9 & 13 \\ \hline
	 Impressionism & 8.5 & 8.6 & 3.7 & 8.6 & 17.2 & 45.7 & 11.4 \\ \hline
	 Cubism & 1.5 & 6.7 & 17.6 & 8.6 & 20 & 0 & 34.3 \\ \hline
    \end{tabular}}
	\end{center}
    \caption{Discriminative BoW using CSIFT}
    \label{tab:DisBoWCSIFT}
\end{table}

\begin{table}
	\begin{center}
	\resizebox{\textwidth}{!}{
    \begin{tabular}{ | l | l | l | l | l | l | l | l |}
    \hline
     Confusion(\%) & Baroque &  Abstract & Renaissance & Pop-Art & Expressionism & Impressionism & Cubism  \\ \hline
     Baroque & 82.1 & 0 & 10.7 & 0 & 14.3 & 17.9 & 3.6 \\ \hline
     Abstract & 0 & 54.2 & 3.6 & 7.1 & 7.1 & 3.6 & 7.1 \\ \hline
	 Renaissance & 3.6 & 0 & 64.3 & 3.6 & 21 & 0 & 7.1 \\ \hline	
	 Pop-Art & 0 & 12.5 & 3.6 & 75 & 0 & 0 & 17.9 \\ \hline
	 Expressionism & 0 & 16.7 & 0 & 3.6 & 36 & 10.7 & 28.6 \\ \hline
	 Impressionism & 14.3 & 8.33 & 7.2 & 3.6 & 10.7 & 57.1 & 7.1 \\ \hline
	 Cubism & 0 & 4.2 & 10.8 & 7.1 & 14.3 & 10.7 & 28.6 \\ \hline
    \end{tabular}}
	\end{center}
    \caption{Discriminative BoW using OSIFT}
    \label{tab:DisBoWOSIFT}
\end{table}

\begin{figure}[htp]
\center
\includegraphics[width=0.9\textwidth,height = 2.5in]{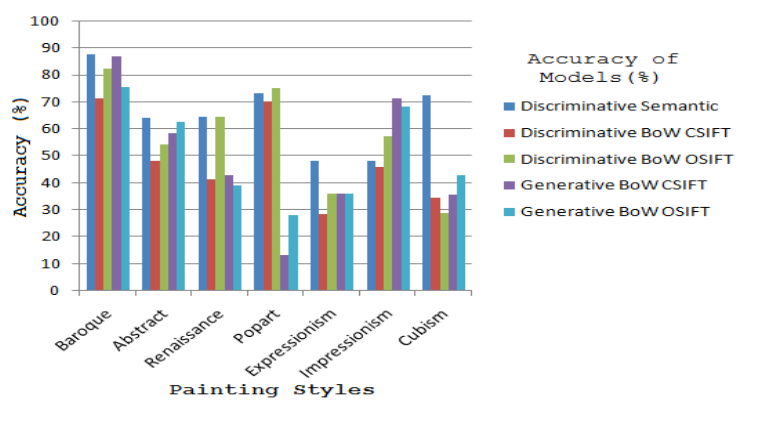}
\caption{Classification accuracy for each approach on each style}
\label{fig:styleclass}
\end{figure}

We evaluated and tested the three models on our dataset, and calculated and compared the classification accuracy for each of them. Table~\ref{tab:DisSemMod} shows the confusion matrix of the Discriminative Semantic Model over the five-fold cross validation. The overall accuracy achieved is 65.4 \%. 
Table~\ref{tab:DisBoWCSIFT} and~\ref{tab:DisBoWOSIFT} show the confusion matrices for the discriminative BoW model with CSIFT and OSIFT features respectively. Overall accuracy achieved is 48.47\% and 56.7\% respectively.  
Table~\ref{tab:GenBoWCSIFT} and~\ref{tab:GenBoWOSIFT} show the confusion matrices for the generative topic model using CSIFT and OSIFT features, with average accuracy of 49\% and 50.3\% respectively. 
Table~\ref{tab:overall} summarizes the overall results for all the experiments. Figure~\ref{fig:styleclass} shows the accuracies for classifying each style using all the evaluated models.

As can be examined from the results, the Discriminative model with Semantic-level features achieved the highest accuracy followed by Discriminative BoW with OSIFT, Generative BoW with OSIFT, Generative BoW with CSIFT and Discriminative BoW CSIFT.
Also it can be deduced from the results that both Discriminative and Generative BoW models achieved comparable accuracy, while Discriminative Semantic model outperforms both BoW models. These results are inline with our hypothesis that the Semantic-level information would be more suitable for the task of fine-art style classification. 
By examining the results we can notice that the Baroque style is always classified with the highest accuracy in all techniques. It is also interesting to notice that the Popart style is classified with accuracy over 70\% in all the discriminative approaches while the generative approach performed poorly in that style. Also it is worth noting that the OSIFT features outperformed the CSIFT features in the discriminative case; however the difference is not significant in the generative case.

\begin{table}
	\begin{center}
	\resizebox{\textwidth}{!}{
    \begin{tabular}{ | l | l | l | l | l | l | l | l |}
    \hline
     Confusion(\%) & Baroque &  Abstract & Renaissance & Pop-Art & Expressionism & Impressionism & Cubism  \\ \hline
     Baroque & 86.6 & 0 & 14.3 & 0 & 14.3 & 7.1 & 7.1 \\ \hline
     Abstract & 0 & 58.3 & 7.1 & 26.6 & 0 & 7.1 & 14.3 \\ \hline
	 Renaissance & 6.6 & 8.3 & 42.8 & 20 & 14.3 & 0 & 7.1 \\ \hline	
	 Pop-Art & 0 & 0 & 7.1 & 13.3 & 0 & 0 & 14.3 \\ \hline
	 Expressionism & 0 & 8.3 & 7.1 & 6.6 & 36 & 14.3 & 7.1 \\ \hline
	 Impressionism & 6.6 & 25 & 14.3 & 13.3 & 21.4 & 71.4 & 14.3 \\ \hline
	 Cubism & 0 & 0 & 7.1 & 20 & 14.3 & 0 & 35.7 \\ \hline
    \end{tabular}}
	\end{center}
    \caption{Generative BoW topic model using CSIFT}
    \label{tab:GenBoWCSIFT}
\end{table}

\begin{table}
	\begin{center}
	\resizebox{\textwidth}{!}{
    \begin{tabular}{ | l | l | l | l | l | l | l | l |}
    \hline
     Confusion(\%) & Baroque &  Abstract & Renaissance & Pop-Art & Expressionism & Impressionism & Cubism  \\ \hline
     Baroque & 75.5 & 0 & 14.3 & 0 & 3.6 & 10.7 & 7.1 \\ \hline
     Abstract & 0 & 62.5 & 3.5 & 27.3 & 3.6 & 3.6 & 0 \\ \hline
	 Renaissance & 7.1 & 4.2 & 39.2 & 3.3 & 7.1 & 3.6 & 10.7 \\ \hline	
	 Pop-Art & 0 & 8.3 & 0 & 28 & 3.6 & 0 & 7.1 \\ \hline
	 Expressionism & 7.1 & 0 & 17.8 & 14 & 36 & 3.6 & 10.7 \\ \hline
	 Impressionism & 10.2 & 25 & 10.7 & 10.2 & 32 & 68 & 21.4 \\ \hline
	 Cubism & 0 & 0 & 14.3 & 16.9 & 14.3 & 10.7 & 42.9 \\ \hline
    \end{tabular}}
	\end{center}
    \caption{Generative BoW topic model using OSIFT}
    \label{tab:GenBoWOSIFT}
\end{table}

\begin{table}
	\begin{center}
	\resizebox{\textwidth}{!}{
    \begin{tabular}{ | l | l | l | l | l | l |}
    \hline
     Model & Dis Semantic & Dis BoW CSIFT & Dis BoW OSIFT & Gen BoW CSIFT & Gen BoW OSIFT\\ \hline	
	 Mean Accuracy(\%) & 65.4 & 48.47 & 56.7 & 49 & 50.3 \\ \hline
	 Std  & 4.8 & 2.45 & 3.26 & 2.43 & 2.46 \\ \hline
	\end{tabular}}
	\end{center}
    \caption{Generative BoW topic model using OSIFT}
    \label{tab:overall}
\end{table}

\section{Influence Discovery Framework}
\label{Sec:Prob_def}
Consider a set of artists,  denoted by $A=\{a^l, l=1\cdots N_a\}$, where $N_a$ is the number of artists. For each artist, $a^l$, we have a set of images of paintings, denoted by $P^l=\{p_i^l, i=1,\cdots,N^l\}$, where $N^l$ is the number of paintings for the $l$-th artist. 
For clarity of the presentation, we reserve the superscript for the artist index and the subscript for the painting index. We denote by $N=\sum_l N_l$ the total number of paintings.
Following the conclusion of the style classification comparative study, we represent each painting by its Classeme features~\cite{aleb}. Therefore, each image $p_i^l \in {R}^D$ is a $D$ dimensional feature vector that is the outcome of the Classeme classifiers, which defines the feature space. 

To represent the temporal information, for each artist we have a  ground truth time period where he/she performed their work, denoted by $t^l= [t_{start}^l, t_{end}^l]$ for the  $l$-th artist, where $t_{start} ^l$ and $t_{end}^l $ are the start and end year of that time period respectively. We do not consider the date of a given painting since for some paintings the exact time is unknown. 

\medskip
\noindent{\bf Painting Similarity:}
\medskip

\begin{figure}[ht]
\centering
\includegraphics[width=.9\textwidth]{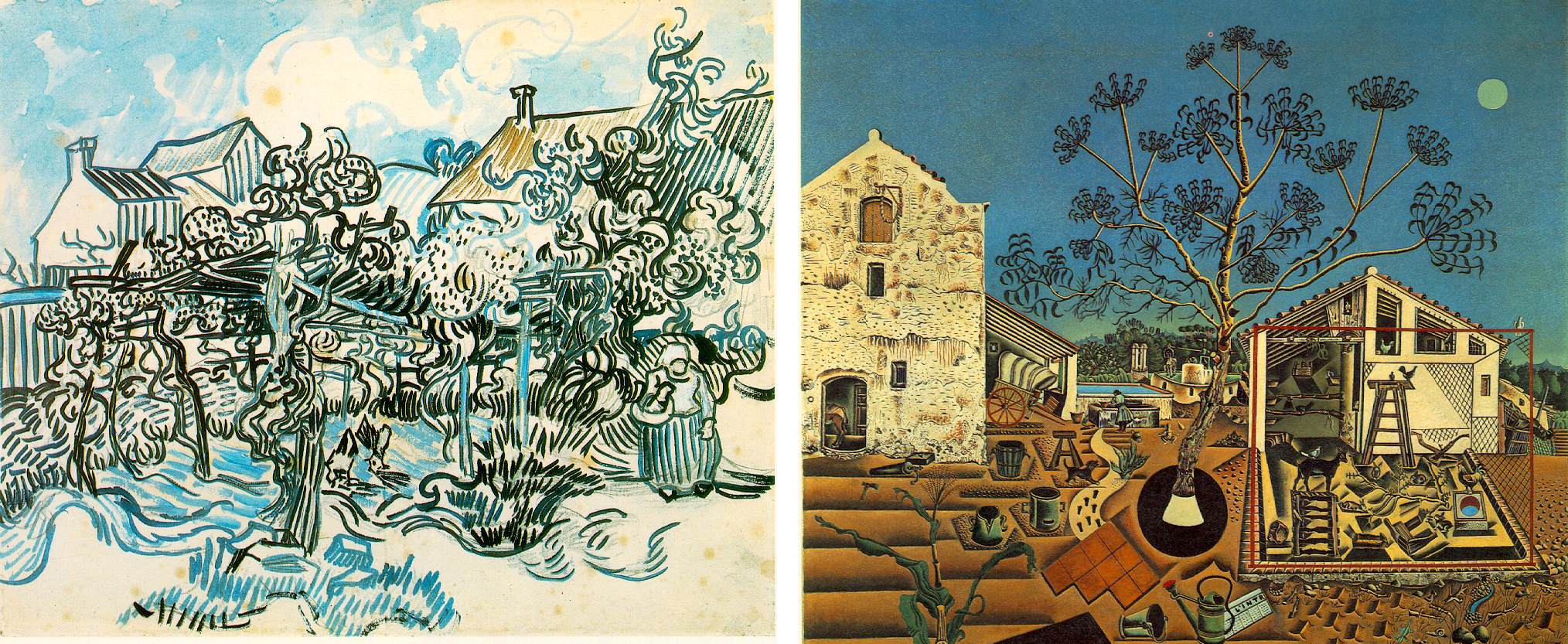}
\caption{Vincent van Gogh's \textit{Old Vineyard with Peasant Woman} 1890 (left) and Joan Miro's \textit{The Farm} 1922 (Right). Similar objects and scenery but different moods and style.}
\label{fig:go}
\end{figure}

\begin{figure}[ht]
\centering
\includegraphics[width=.9\textwidth]{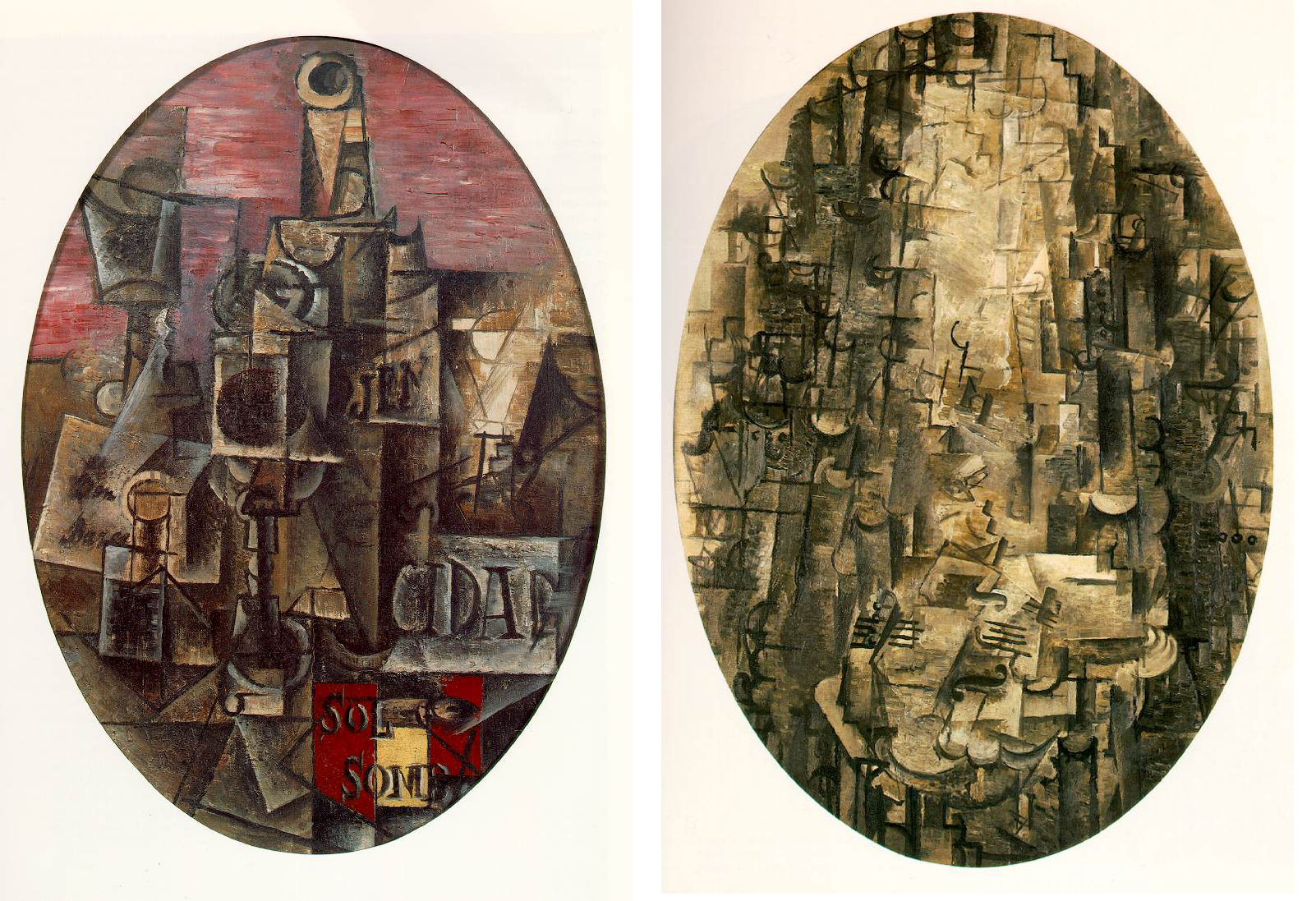}
\caption{Georges Braque's \textit{Man with a Violin} 1912 (Left) and Pablo Picasso's \textit{Spanish Still Life: Sun and Shadow} 1912 (Right).}
\label{fig:braque}
\end{figure}

To encode similarity/dissimilarity between paintings, we consider two different distances:

\medskip
{\em Euclidean distance:} The distance $d_E(p_i^l,p_j^k) $ is defined to be the Euclidean distance between the Classeme feature vectors of paintings $p_i^l$ and $p_j^k$. Since Classeme features are high-level semantic features, the Euclidean distance in the feature space is expected to measure dissimilarity in the subject matter between paintings.  Painting similarity based on the Classeme features showed some interesting cases, several of which have not been studied before by art historians as a potential comparison. Figure~\ref{fig:figure2} is an example of this, as well as Figure~\ref{fig:go} and Figure~\ref{fig:braque}.

\medskip
{\em Manifold distance:} Since the paintings in the feature space are expected to lie on a low-dimensional manifold, the Euclidean distance might be misleading in judging similarity/dissimilarity. Therefore, we also consider a manifold-based distance,  $d_M(p_i^l,p_j^k) $ denoting the geodesic distance along the  manifold of paintings in the feature space. To define such a distance, we use a method similar to ISOMAP~\cite{ISOMAP}, where we build a k-nearest neighbor graph of paintings, and compute the shortest path between each pair of paintings $p_i^l$ and $p_j^k$ on that graph. The distance $d_M(p_i^l,p_j^k) $ is then defined as the sum of the distances along the shortest path. 

\medskip
\noindent{\bf Artist Similarity:}
\medskip

Once painting similarity is encoded, using any of the two methods mentioned above, we can design a suitable similarity measure between artist. There are two challenges to achieve this task. First, how to define a measure of similarity between two artists, given their sets of paintings. We need to define a proper set distance $D(P^l,P^k)$ to encode the distance between the work of the $l$-th and $k$-th artists. This relates to how to define influence between artists to start with, where there is no clear definition.  Should we declare an influence if one paining of artist $k$ has strong similarity to a painting of artist $l$ ? or if a number of paintings have similarity ? and what that ``number'' should be ?

Mathematically speaking, for a given painting  $ p_i^l \in P^l$ we can find its closest painting in $P^k$ using a point-set distance as 
\[  d(p_i^l, P^k ) = \min_j  d(p_i^l,p_j^k). \]
 We can find one painting in by artist $l$ that is very similar to a painting by artist $k$, that can be considered an influence. This dictates defining an asymmetric distance measure in the form of 
\[ D_{min}(P^l,P^k) = \min_i d(p_i^l, P^k ). \]  
We denote this measure by {\em minimum-link influence}.

  On the other hand, we can consider a central tendency in measuring influence, where we can measure the average or median of painting distances between $P^l$ and $P^k$, we denote this measure {\em central-link influence}.

Alternatively, we can think of Hausdorff distance~\cite{dubuisson}, which measures the distance between two sets as the supremum of the point-set distances, defined as
\[  D_H(P^l,P^k) = \max (\max_i d(p_i^l, P^k ) , \max_j d(p_j^k, P^l )   ) . \]
We denote this measure {\em maximum-link influence}.
Hausdorff distance is widely used in matching spatial points, which unlike a minimum distance, captures the configuration of all the points.  While the intuition of Hausdorff distance is clear from a geometrical point of view, it is not clear what it means in the context of artist influence, where each point represent a painting.
In this context, Hausdorff distance measures the maximum distance between any painting and its closest painting in the other set. 

The discussion above highlights the challenge in defining the similarity between artists, where each of the suggested distance is in fact meaningful, and captures some aspects of similarity, and hence influence. In this paper, we do not take a position in favor of any of these measures, instead we propose to use a measure that can vary through the whole spectrum of distances between two sets of paintings. We define asymmetric distance between artist $l$ and artist $k$ as the $q$-percentile Hausdorff distance, as  
\begin{equation}
\label{Eq:percentile}
 D_{q\%}(P^l,P^k) =  \max_{i}^{q\%} d(p_i^l, P^k )    . 
 \end{equation}
Varying the percentile $q$ allows us to evaluate different settings ranging from a minimum distance, $D_{min}$, to a central tendency, to a maximum distance as in Hausdorff distance $D_H$. 

\medskip
\noindent{\bf  Artist Influence Graph:}
\medskip

The artist asymmetric distance is used, in conjunction with the ground-truth time period to construct an influenced-by graph. The influence graph is a directed graph where each artist is represented by a node. A weighted directed edge between node $i$ and node $j$ indicates that artist $i$ is potentially influenced by artist $j$, which is only possible if artist $i$ succeed or is contemporary to artist $j$.    
The weight corresponds to the artist distance, i.e., a smaller weight indicates a higher potential influence.  Therefore, the graph weights are defined as
\begin{equation}
 w_{ij} = \left\{ \begin{array}{lll}  D_{q\%}(P^i,P^j) & \hspace{1cm} &  \mbox{if} \;\;  t_{end} ^i  \geq  t_{start}^j \\
 \infty & & \mbox{otherwise} \\ \end{array} \right.
\end{equation}

\section{Influence Discovery Results}
\label{Sec:Exp}
\label{Sec:influence_disc}
\subsection{Evaluation Methodology:}

\begin{figure}[htbp]
\centering
\includegraphics[height=7in]{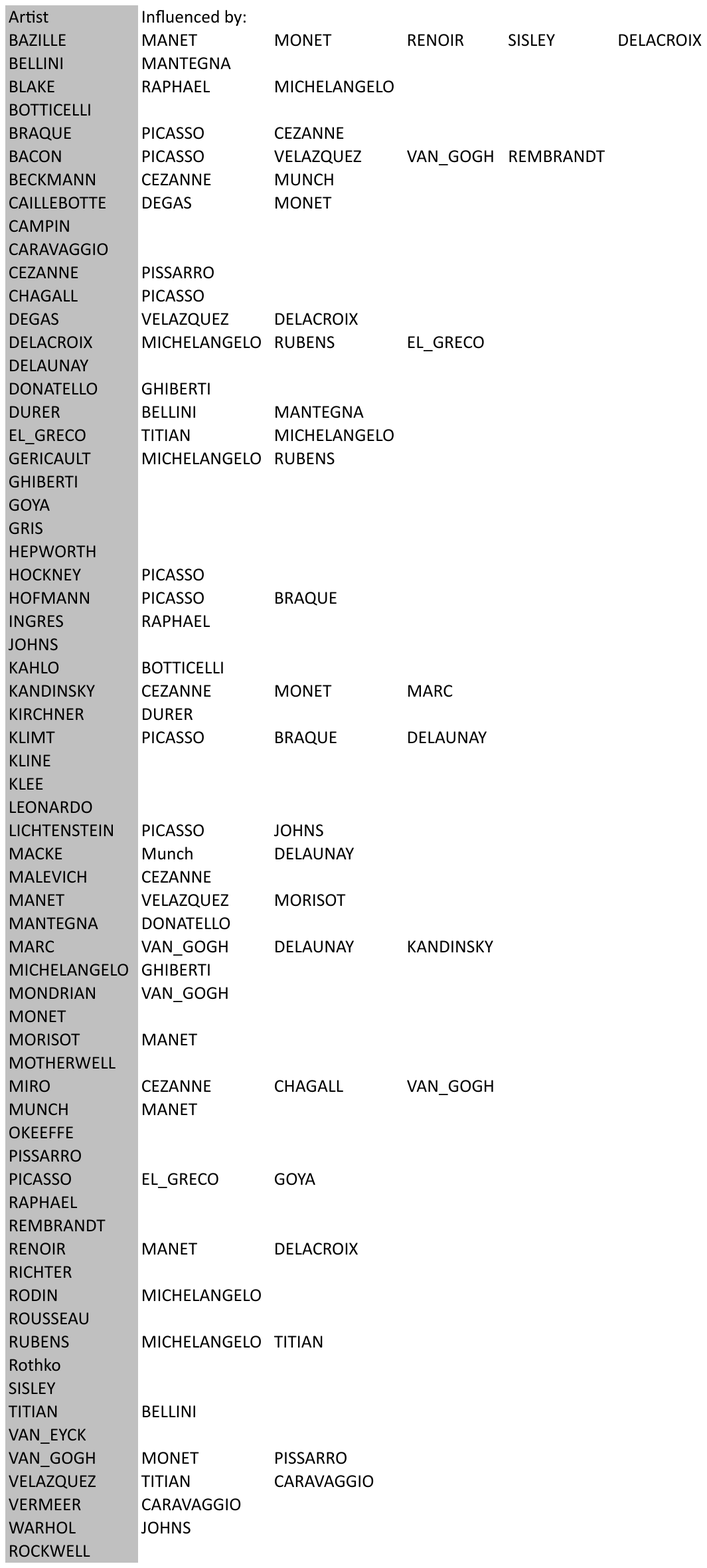}
\caption{Ground-truth influences}
\label{Influencedbylist}
\end{figure}

We researched known influences between artists within our dataset from multiple resources such as \textit{The Art Story Foundation} and \textit{The Metropolitan Museum of Art}. For example, there is a general consensus among art historians that Paul C\'ezanne's use of fragmented spaces had a large impact on Pablo Picasso's work.  In total, we collected 76 pairs of one-directional artist influences, where a pair $(a^i, a^j)$ indicates that artist $i$ is influenced by artist $j$. Figure~\ref{Influencedbylist} shows the complete list of influenced-by list. Generally, it is a sparse list that contains only the influences which are consensual among many. Some artists do not have any influences in our collection while others may have up to five. We use this list as ground-truth for measuring the accuracy in our experiments. 

The constructed influenced-by graph is used to retrieve the top-k potential influences for each artist. If a retrieved influence pair concur with an influence ground-truth pair, this is considered a hit.  The hits are used to compute the recall, which is defined as  the ratio between the correct influence detected and the total known influences in the ground truth. The recall is used for the sake of comparing the different settings relatively. Since detected influences can be correct although not in our ground truth, so there is no meaning to compute the precision.

\subsection{Influence Discovery Validation}

\begin{figure}  
\centering
\begin{minipage}{\textwidth}  
    \includegraphics[width=.5\textwidth]{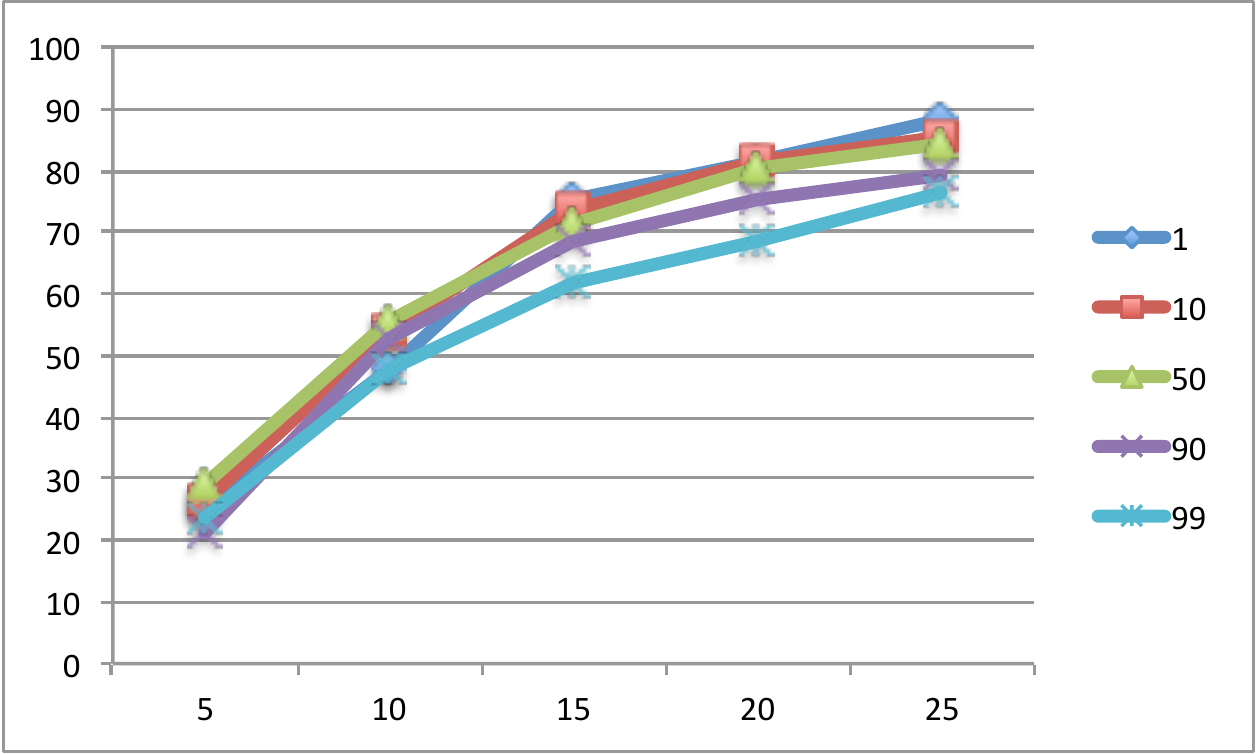}
    \includegraphics[width=.5\textwidth]{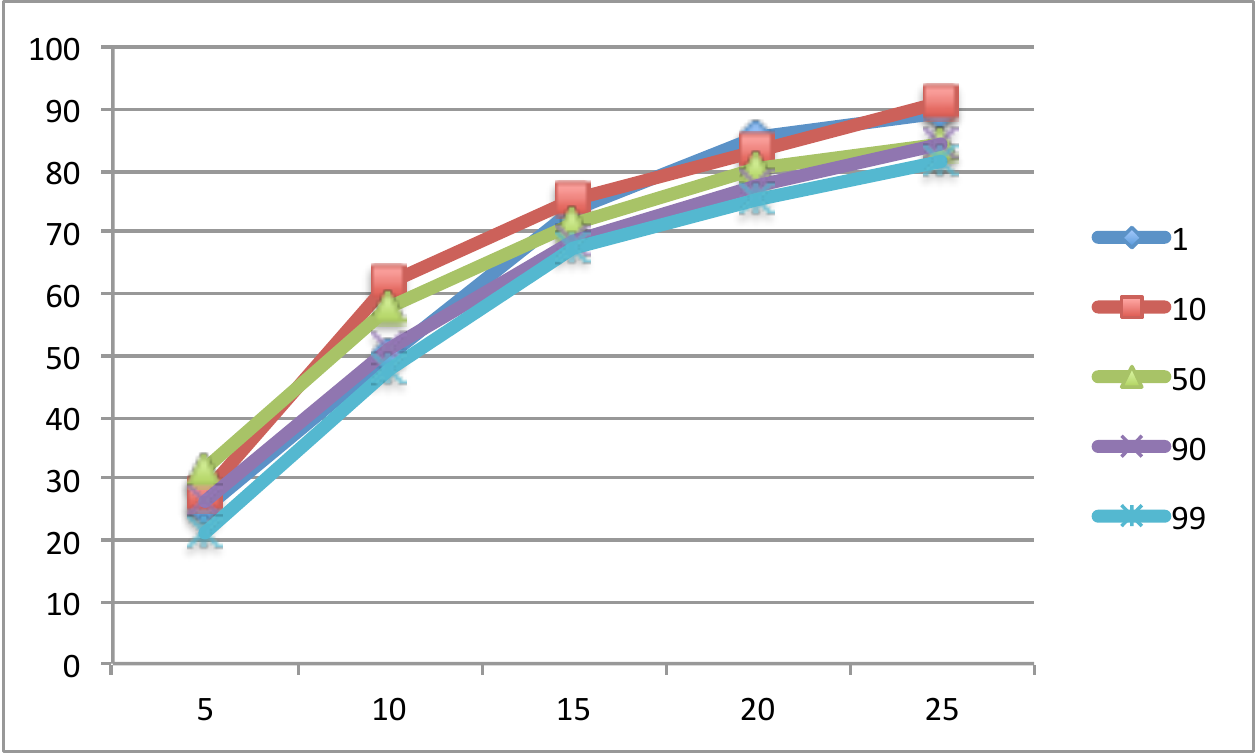}
  \caption{Influence recall curves, using classemes features with different q\%. Left: Euclidean distance, Right: Manifold distance. }
  \label{fig:recall1}
  \end{minipage}
\end{figure}
\begin{figure}  
\centering
\begin{minipage}{1\textwidth}  
    \includegraphics[width=0.5\textwidth]{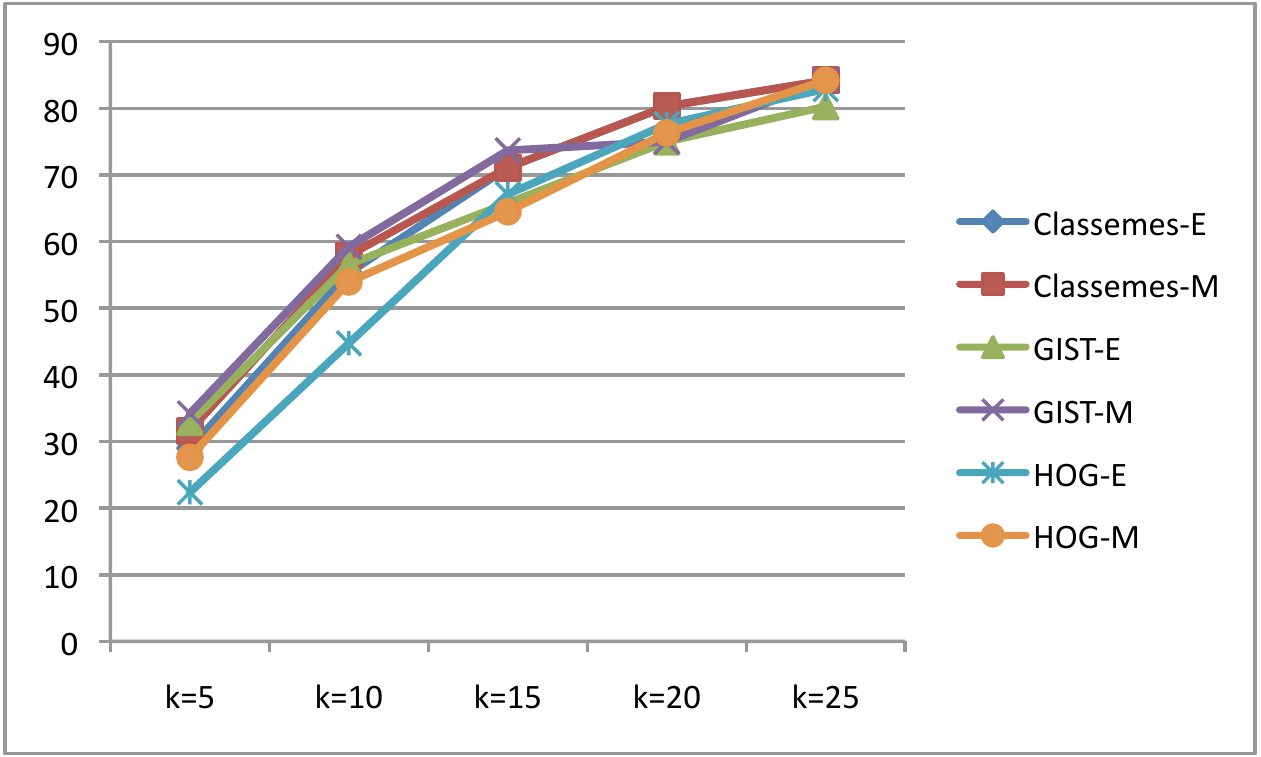}
    \includegraphics[width=0.5\textwidth]{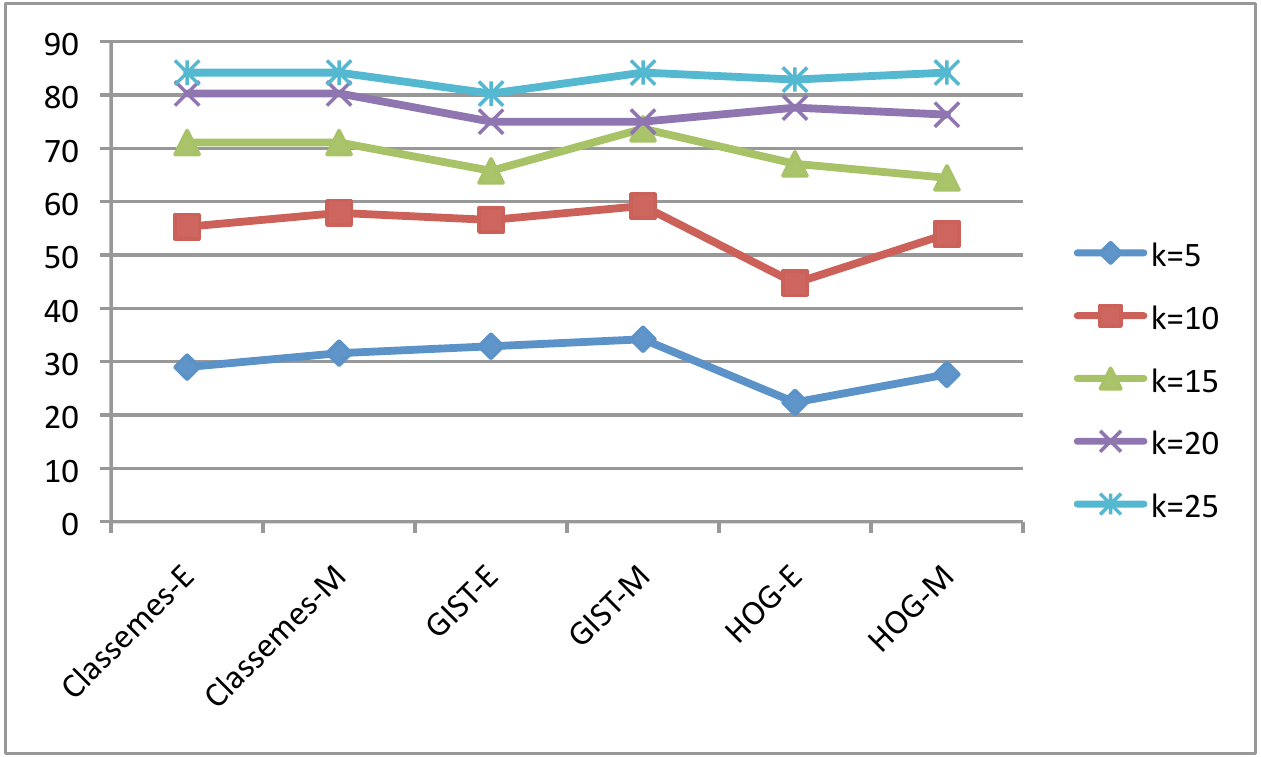}
  \caption{Influence Recall at different top-k: Comparisons of different descriptors}
  \label{fig:recall2}
  \end{minipage}
\end{figure}

We experimented with the Classeme features, which showed the best results in the style classification task. We also experimented with GIST descriptors~\cite{GIST} and HOG descriptors~\cite{HOG}, since they are the main ingredients in the Classemes features. In all cases, we computed the recall figures using the influence graph for the top-k similar artist (k=5, 10, 15, 25) with different $q$-percentile for the artist distance measure in Eq~\ref{Eq:percentile} (q=1, 10, 50, 90, 99\%). For all descriptors, we computed the influences using both the Euclidean distance and the Manifold-based distances. The results are shown in Tables~\ref{tab:euclidean}-~\ref{tab:hog_manifold}. The rows of the tables show different $q$-percentile. The columns show the recall percentage for the top-k similar artists . 
From the difference results we can see that most of the time the 50\%-set distance (central-link influence) gives better results. We can also notice that generally the manifold-based distance slightly out performs the Euclidean distance for the same feature. 
Figure~\ref{fig:recall1} shows the recall curves using the Classemes features with different $q\%$. Figure~\ref{fig:recall2} compares the recall curves for different features (Classemes, GIST, HOG) and distances (Euclidean vs Manifolds), all calculated using the 50\% set distance. The results using the three features seems to be comparable. 


\begin{table}
\begin{center}
{
\centering
    \caption{Performance of influence retrieval using Euclidean distance and Classemes features.}
    \begin{tabular}{ | l | l | l | l | l | l |}
    \hline
    & \multicolumn{5}{c |}{top-k recall} \\ \hline
  q\%   &	5&	10&	15&	20&	25 \\ \hline \hline
1&	25&	47.4	& 75&	81.6	&88.2\\ \hline
10&	26.3	&54&	73.7	&81.6	&85.5\\ \hline
50&	29&	55.3&	71.1&	80.3&	84.2\\ \hline
90&	21.1	&52.6&	68.4	&75&	79\\ \hline
99&	23.7&	47.4&	61.8&	68.4&	76.3\\ \hline
    \end{tabular}
}
	\end{center}
    \label{tab:euclidean}
\end{table}

\begin{table}
	\begin{center}
	    \caption{Performance of influence retrieval using manifold-based distance and Classemes features.}
    \begin{tabular}{ | l | l | l | l | l | l |}
    \hline
        & \multicolumn{5}{c |}{top-k recall} \\ \hline
q\%&	5&	10&	15&	20&	25\\ \hline \hline
1&	25&	50&	73.7&	85.5&	89.5\\ \hline
10&	27.6	&61.8	&75&	83&	90.8\\ \hline
50&	31.6	&57.9&	71.1&	80.3&	84.2\\ \hline
90&	26.3	&51.3	&68.4	&77.6	&84.2\\ \hline
99&	21.1&	47.4&	67.1&	75&	81.6\\ \hline
    \end{tabular}
	\end{center}
    \label{tab:manifold}
\end{table}

\begin{table}
	\begin{center}
	    \caption{Performance of influence retrieval using Euclidean distance and GIST features.}
    \begin{tabular}{ | l | l | l | l | l | l |}
    \hline
        & \multicolumn{5}{c |}{top-k recall} \\ \hline
 q\%   &	5&	10&	15&	20&	25\\ \hline \hline
1&21.05&	40.79&	60.53&	69.74&	75.00 \\ \hline
10&31.58	&50.00&	65.79&	71.05&	76.32 \\ \hline
50&32.89	&56.58&	65.79&	75.00&	80.26 \\ \hline
90&28.95&	55.26&	72.37&	76.32&	84.21 \\ \hline
99&23.68&	48.68&	68.42&	76.32&	81.58 \\ \hline
    \end{tabular}
	\end{center}
    \label{tab:gist_euclidean}
\end{table}

\begin{table}
	\begin{center}
   \caption{Performance of influence retrieval using manifold-based distance and GIST features.}
    \begin{tabular}{ | l | l | l | l | l | l |}
    \hline
        & \multicolumn{5}{c |}{top-k recall} \\ \hline
 q\%   &	5&	10&	15&	20&	25\\ \hline \hline
1& 22.37& 	42.11&	63.16&	68.42&	73.68 \\ \hline
10&34.21&	53.95&	67.11&	69.74&	78.95 \\ \hline
50&34.21&	59.21&	73.68&	75.00&	84.21 \\ \hline
90&30.26&	55.26&	71.05&	73.68&	78.95 \\ \hline
99 & 21.05 &	48.68 &	67.11 &	73.68 &	81.58 \\ \hline
    \end{tabular}
	\end{center}
    \label{tab:gist_manifold}
\end{table}

\begin{table}
	\begin{center}
    \caption{Performance of influence retrieval using Euclidean distance and HOG features.}
    \begin{tabular}{ | l | l | l | l | l | l |}
    \hline
        & \multicolumn{5}{c |}{top-k recall} \\ \hline
  q\%  &	5&	10&	15&	20&	25\\ \hline \hline
1&22.37&	40.79&	56.58&	71.05&	78.95 \\ \hline
10&22.37	&47.37&	64.47&	78.95&	82.89 \\ \hline
50&22.37	&44.74&	67.11&	77.63&	82.89 \\ \hline
90&25.00&	52.63&	67.11&	77.63&	84.21 \\ \hline
99 & 26.32 &	48.68 &	63.16 &	73.68 &	78.95 \\ \hline
    \end{tabular}
	\end{center}
    \label{tab:hog_euclidean}
\end{table}

\begin{table}
	\begin{center}
    \caption{Performance of influence retrieval using manifold-based distance and HOG features.}
    \begin{tabular}{ | l | l | l | l | l | l |}
    \hline
        & \multicolumn{5}{c |}{top-k recall} \\ \hline \hline
 q\%   &	5&	10&	15&	20&	25\\ \hline
1&23.68&	39.47&	57.89&	71.05&	80.26 \\ \hline
10&25.00&	46.05&	63.16&	76.32&	80.26 \\ \hline
50&27.63&	53.95&	64.47&	76.32&	84.21 \\ \hline
90&23.68&	46.05&	65.79&	75.00&	81.58 \\ \hline
99 & 27.63 &	43.42 &	57.89 &	68.42 &	71.05 \\ \hline
    \end{tabular}
	\end{center}
    \label{tab:hog_manifold}
\end{table}

\subsection{Visualizing Influences - A Map of Artists}

The influence graph can be used to achieve a visualization of artists and their similarities, i.e. a Map of Artists. For this purpose we used ISOMAP~\cite{ISOMAP} to achieve a low-dimensional embedding of the artist influence graph. ISOMAP computes the shortest path on the graph between each two artists, and use that to achieve an embedding using multi-dimensional scaling (MDS)~\cite{BorgGroenen2005}. The reason we use ISOMAP in particular, among several other low-dimensional embedding techniques, is that ISOMAP works with directed graphs.

%

\begin{figure}[htp]
\includegraphics[width=\textwidth,height = 3.5in]{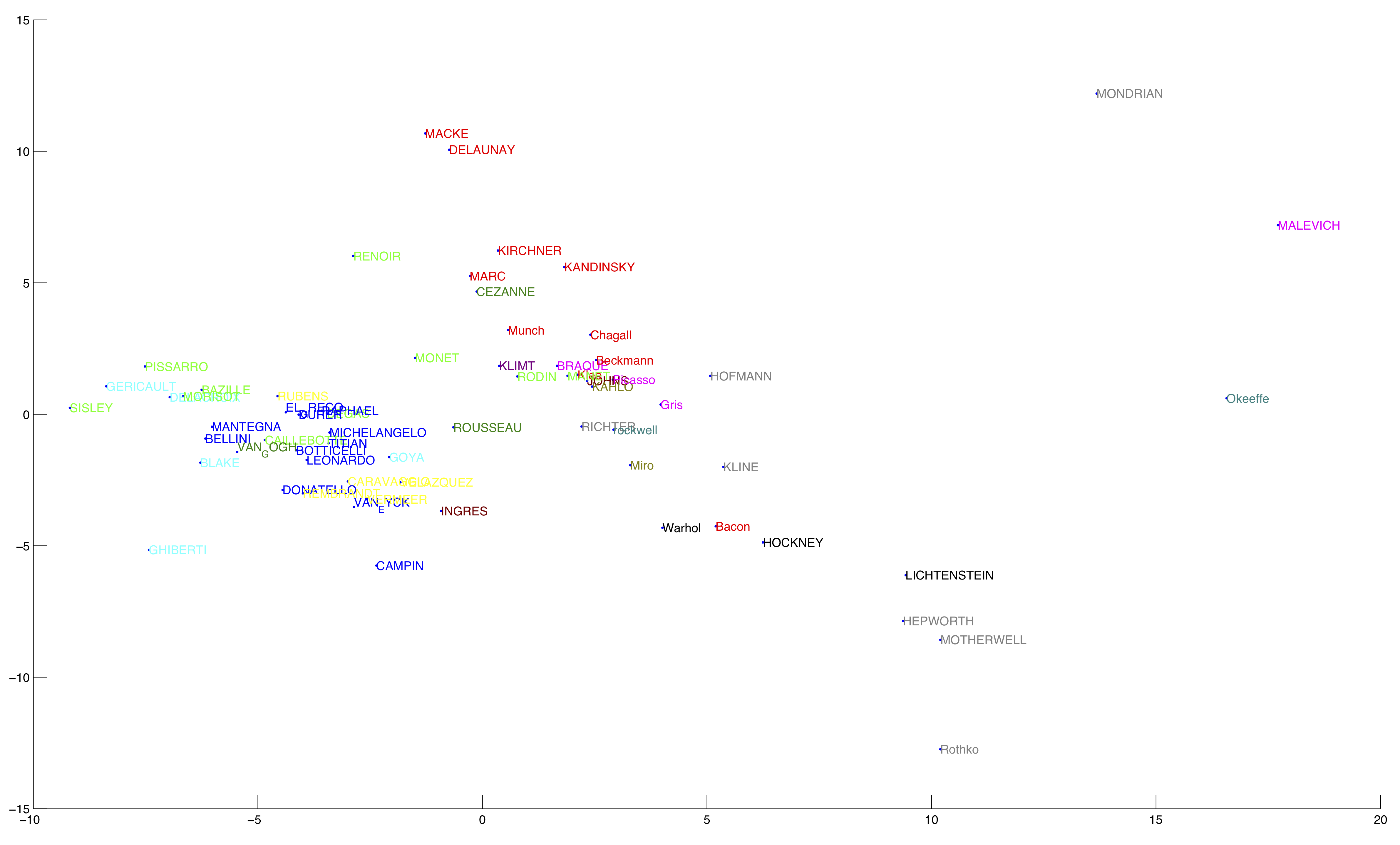}\\
\includegraphics[width=\textwidth,height = 3.5in]{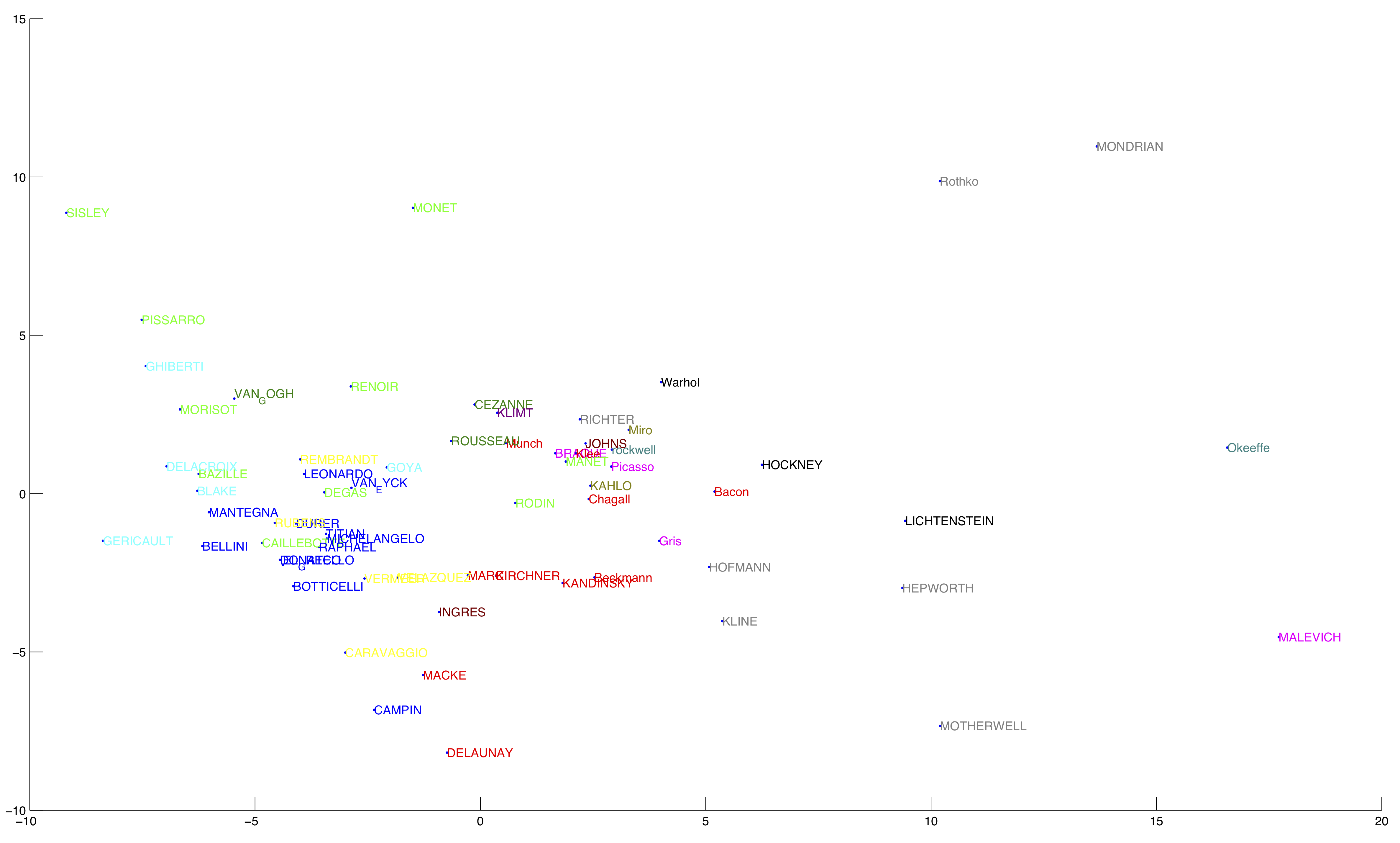}
\caption{Map of Artists: Similar artists in two dimensions: Top: Dimensions 1 and 2, Bottom: Dimensions 1 and 3. Artist are color coded by their style.}
\label{fig:isomap2D_1}
\end{figure}

\begin{figure}[htp]
\includegraphics[width=\textwidth,height = 3.5in]{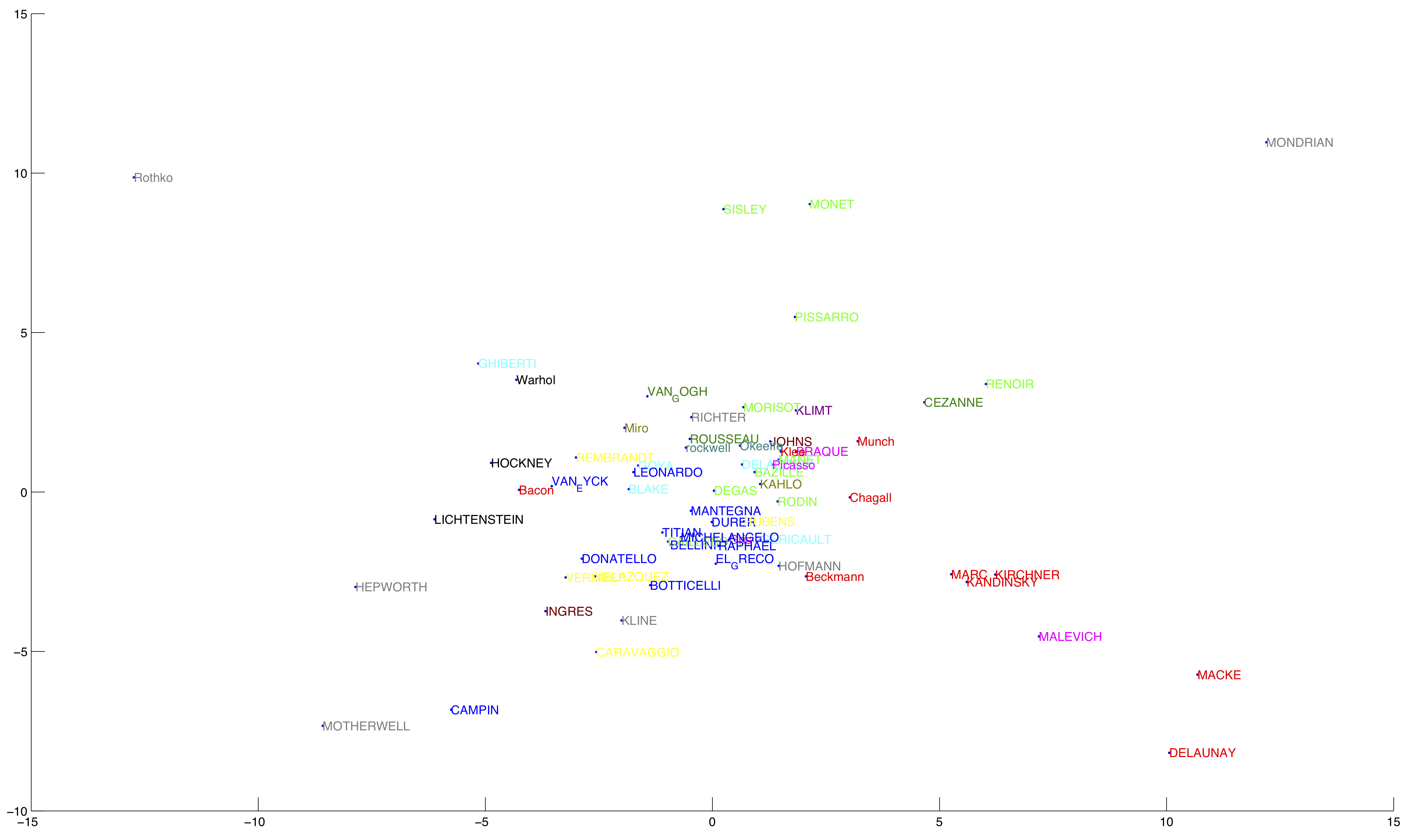}\\
\caption{Map of Artists: Similar artists in two dimensions: Top: Dimensions 2 and 3. Artists are color coded by their style.}
\label{fig:isomap2D_2}
\end{figure}

%

Figure~\ref{fig:isomap2D_1} and~\ref{fig:isomap2D_2} illustrate a visualization of artist similarity based on embedding the influence graph into a three-dimensional space using ISOMAP, each plot shows a two-dimensional projection of that space. The artists are color coded in these plots to reflect their ground-truth style. 
We can see that artist of the same style are mostly clustered together. For example a few \textit{Expressionist} artists  clustered together as well as \textit{Abstract Contemporary} artists. As seen, the artists populated the right of the mapping are Lichtenstein, Hepworth, Malevich, Mondrian, Motherwell, O'Keffe, and Rothko, who are all Modern and Abstract artists. Their styles differ slightly but all share some stylistic approaches and time period. On the left side of the plot we can find most  \textit{Impressionists} and \textit{Renaissance} artists. However, we can see that the \textit{Impressionists} and \textit{Renaissance} artists seem to have similar values in one dimension but not the other. It is also clear that the distances within and between the \textit{Impressionists} and \textit{Renaissance} (in the right side) are much smaller than the distances among the \textit{Expressionist} and \textit{Abstract Contemporary} artists (in the left side).  Other styles, such as \textit{Romanticism}, seem to have a broader range of values. 

\begin{figure}[htbp]
\centering
\includegraphics[height=7in]{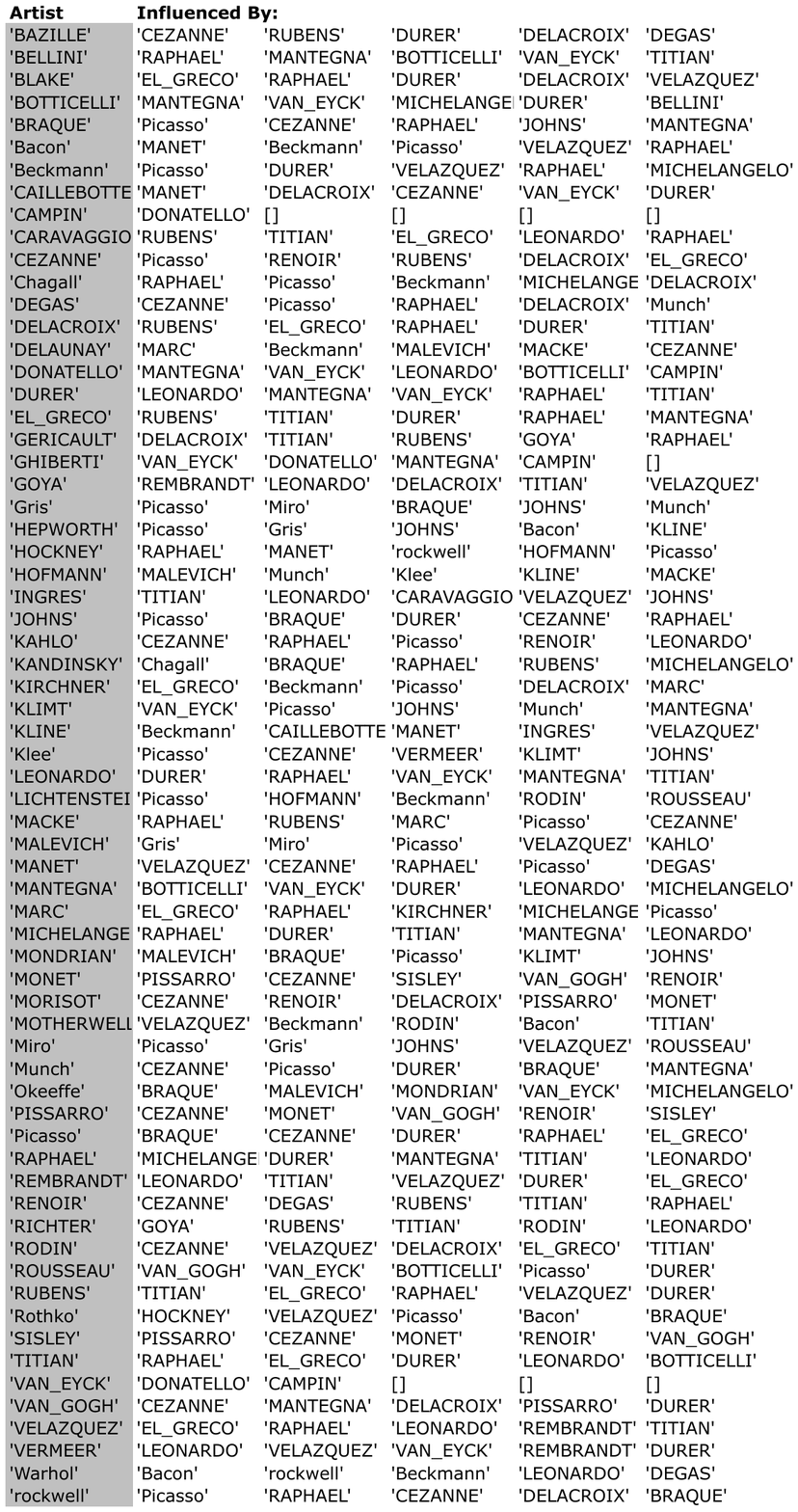}
\caption{Top-5 suggested influences retrieved from the graph: using Classemes features, Euclidean distance, and q=50\%, }
\label{ResInfluencedbylist}
\end{figure}

Some artists in this mapping seem to cluster according to their style, but in the context of influence, it is also important to think about the similarities between artists instead of the classification of style.  This is yet another complication of the task of measuring influence.
Therefore, another way to analyze this graph is to disregard style all together. We can wonder whether Richter and Hockney share a connection because they lie close to each other. Or we can wonder if Klimt was influenced by Picasso or Braque. In fact, both Picasso and Braque were listed as influences for Klimt in our ground-truth list. When comparing these close mappings to the ground truth influence, some are reasonable while others seem less coherent. In another example, Bazille lies close to Delacroix which is consistent with our ground truth. 
Other successful mappings include Munch's influence on Beckmann,  Degas's influence on Caillebotte, and others.
Figure~\ref{ResInfluencedbylist} illustrates the top-5 suggested influence results.  

\section{Conclusion and Future Works}
\label{Sec:con}

This paper scratches the surface of the problem of automated discovery of artist influence, through the study of painting and artist similarity.  We posed the interesting question of finding influence between painters as a knowledge discovery problem and showed interesting results for both of the qualitative and quantitative measurements.  

In this paper we also studied the problem of paintings style classification, and presented a comparative study of three different models for the classification task, with different visual features. That study showed that semantic-level features perform the best for this task. This conclusion lead us to use these semantic features for the task of influence discovery. 

For the task of influence discovery, we compared several distance measures between paintings, including a Euclidean distance and a manifold-based distance. The comparative experiments showed that the manifold-based distance gave slightly better results. We proposed and evaluated different artist distance measures, denoted as minimum-link, central-link, and maximum-link influence measures. This problem can be formulated as a set distance, however the typical Hausdorff set distance did not perform best, instead the central-link influence measure performed best in all experiments. We also present a tool for visualizing artist similarity through what we call a map of artists. 

In this paper we also presented a new annotated dataset with diverse set of artists and wide range of paintings. This dataset will be publicly available and can be used for interdisciplinary tasks of Art and Computer Science. 

Of course, there is a lot more to be done. For example, our framework could include searching for specific stylistic similarities such as brushstroke and pattern. We could also include more features of color and line. We can experiment with many other features especially among the elements and principles of art. Clearly there are many ways in which artists are influenced by each other. This is why mapping influence is such a difficult task.


%
%


{\small
\bibliographystyle{plain}
\bibliography{egbib}

\begin{thebibliography}{10}

\bibitem{CSIFT}
A.~E. Abdel-Hakim and A.~A. Farag.
\newblock Csift: A sift descriptor with color invariant characteristics.
\newblock In {\em IEEE Conference on Computer Vision and Pattern Recognition,
  CVPR}, 2006.

\bibitem{Arora12}
Ravneet~Singh Arora and Ahmed~M. Elgammal.
\newblock Towards automated classification of fine-art painting style: A
  comparative study.
\newblock In {\em ICPR}, 2012.

\bibitem{Belkin02laplacianeigenmaps}
Mikhail Belkin and Partha Niyogi.
\newblock Laplacian eigenmaps for dimensionality reduction and data
  representation.
\newblock {\em Neural Computation}, 15:1373--1396, 2002.

\bibitem{BorgGroenen2005}
I.~Borg and P.J.F. Groenen.
\newblock {\em {Modern Multidimensional Scaling: Theory and Applications}}.
\newblock Springer, 2005.

\bibitem{SVM}
Christopher J.~C. Burges.
\newblock A tutorial on support vector machines for pattern recognition.
\newblock {\em Data Mining and Knowledge Discovery}, 2:121--167, 1998.

\bibitem{CabralCDBC11}
Ricardo~S. Cabral, JoÃ£o~P. Costeira, Fernando {De la Torre}, Alexandre
  Bernardino, and Gustavo Carneiro.
\newblock Time and order estimation of paintings based on visual features and
  expert priors.
\newblock In {\em SPIE Electronic Imaging, Computer Vision and Image Analysis
  of Art II}, 2011.

\bibitem{Carneiro11}
Gustavo Carneiro.
\newblock Graph-based methods for the automatic annotation and retrieval of art
  prints.
\newblock In {\em ICMR}, 2011.

\bibitem{Carneiro12}
Gustavo Carneiro, Nuno~Pinho da~Silva, Alessio~Del Bue, and Jo{\~a}o~Paulo
  Costeira.
\newblock Artistic image classification: An analysis on the printart database.
\newblock In {\em ECCV}, 2012.

\bibitem{libsvm}
Chih-Chung Chang and Chih-Jen Lin.
\newblock {LIBSVM}: A library for support vector machines.
\newblock {\em ACM Transactions on Intelligent Systems and Technology},
  2:27:1--27:27, 2011.

\bibitem{blei}
A.~Ng D.~Blei and M.~Jordan.
\newblock Latent dirichlet allocation.
\newblock In {\em Journalof Machine Learning Research}, 2003.

\bibitem{HOG}
Navneet Dalal and Bill Triggs.
\newblock Histograms of oriented gradients for human detection.
\newblock In {\em International Conference on Computer Vision \& Pattern
  Recognition}, volume~2, pages 886--893, June 2005.

\bibitem{dubuisson}
M-P Dubuisson and Anil~K Jain.
\newblock A modified hausdorff distance for object matching.
\newblock In {\em Pattern Recognition}, 1994.

\bibitem{khan}
Maria~Vanrell Fahad Shahbaz~Khan, Joost van de~Weijer.
\newblock Who painted this painting?
\newblock 2010.

\bibitem{fei}
Li~Fei-fei.
\newblock A bayesian hierarchical model for learning natural scene categories.
\newblock In {\em In CVPR}, 2005.

\bibitem{lois}
Lois Fichner-Rathus.
\newblock {\em Foundations of Art and Design}.
\newblock Clark Baxter.

\bibitem{Csurka2004}
L.X. Fan J.~Willamowski G.~Csurka, C.~Dance and C.~Bray.
\newblock Visual categorization with bags of keypoints.
\newblock In {\em Proc. of ECCV International Workshop on Statistical Learning
  in Computer Vision}, 2004.

\bibitem{Graham10}
Friedenberg J. Rockmore~D. Graham, D.
\newblock Mapping the similarity space of paintings: image statistics and
  visual perception.
\newblock {\em Visual Cognition}, 2010.

\bibitem{brdahujapo09}
Andrei Brasoveanu Shannon Hughes Ingrid~Daubechies Gungor~Polatkan,
  Sina~Jafarpour.

\bibitem{artchive}
Mark Harden.
\newblock The artchive{@http://artchive.com/cdrom.htm}.

\bibitem{widjaja}
W.~Leow I.~Widjaja and F.~Wu.
\newblock Identifying painters from color profiles of skin patches in painting
  images.
\newblock In {\em ICIP}, 2003.

\bibitem{sivic}
Alexei A. Efros Andrew Zisserman William T.~Freeman Josef~Sivic, Bryan
  C.~Russell.
\newblock Discovering objects and their location in images.
\newblock In {\em ICCV}, 2005.

\bibitem{osift}
Theo~Gevers Koen E. A. van~de Sande and Cees G.~M. Snoek.
\newblock Evaluating color descriptors for object and scene recognition.
\newblock In {\em IEEE Transactions on Pattern Analysis and Machine
  Intelligence}, 2010.

\bibitem{Jia12}
Jia Li, Lei Yao, Ella Hendriks, and James~Z. Wang.
\newblock Rhythmic brushstrokes distinguish van gogh from his contemporaries:
  Findings via automated brushstroke extraction.
\newblock {\em IEEE Trans. Pattern Anal. Mach. Intell.}, 2012.

\bibitem{Lombardi}
Thomas~Edward Lombardi.
\newblock The classification of style in fine-art painting.
\newblock {\em ETD Collection for Pace University. Paper AAI3189084.}, 2005.

\bibitem{SIFT}
David~G. Lowe.
\newblock Distinctive image features from scale-invariant keypoints.
\newblock {\em Int. J. Comput. Vision}, 2004.

\bibitem{Ng2001}
Andrew~Y. Ng and Michael~I. Jordan.
\newblock On discriminative vs. generative classifiers: A comparison of
  logistic regression and naive bayes, 2001.

\bibitem{GIST}
Aude Oliva and Antonio Torralba.
\newblock Modeling the shape of the scene: A holistic representation of the
  spatial envelope.
\newblock {\em International Journal of Computer Vision}, 42:145--175, 2001.

\bibitem{sablatnig}
P.~Kammerer R.~Sablatnig and E.~Zolda.
\newblock Hierarchical classification of paintings using face- and brush stroke
  models.
\newblock In {\em ICPR}, 1998.

\bibitem{robert}
Robert {Sablatnig}, Paul {Kammerer}, and Ernestine {Zolda}.
\newblock Structural analysis of paintings based on brush strokes.
\newblock In {\em Proc. of SPIE Scientific Detection of Fakery in Art}. SPIE,
  1998.

\bibitem{ShiHD09}
Fanhuai Shi, Xixia Huang, and Ye~Duan.
\newblock Robust harris-laplace detector by scale multiplication.
\newblock In {\em ISVC (1) Lecture Notes in Computer Science}.

\bibitem{BoW}
Josef Sivic and Andrew Zisserman.
\newblock Efficient visual search of videos cast as text retrieval.
\newblock {\em IEEE Trans. Pattern Anal. Mach. Intell.}

\bibitem{ISOMAP}
J.~B. Tenenbaum, V.~Silva, and J.~C. Langford.
\newblock {A Global Geometric Framework for Nonlinear Dimensionality
  Reduction}.
\newblock {\em Science}, 290(5500):2319--2323, 2000.

\bibitem{BoW3D}
Roberto Toldo, Umberto Castellani, and Andrea Fusiello.
\newblock A bag of words approach for 3d object categorization.
\newblock In {\em Proceedings of the 4th International Conference on Computer
  Vision/Computer Graphics CollaborationTechniques}, 2009.

\bibitem{aleb}
Lorenzo Torresani, Martin Szummer, and Andrew Fitzgibbon.
\newblock Efficient object category recognition using classemes.
\newblock In {\em ECCV}, 2010.

\bibitem{VanDeSande2010}
Theo van~de Sande, Koen;~Gevers and Cees G.~M. Jan-Snoek.
\newblock Evaluating color descriptors for object and scene recognition.
\newblock {\em IEEE Transactions on Pattern Analysis and Machine Intelligence},
  32(9), 2010.

\bibitem{weinberger2004learning}
Kilian~Q Weinberger, Fei Sha, and Lawrence~K Saul.
\newblock Learning a kernel matrix for nonlinear dimensionality reduction.
\newblock In {\em Proceedings of the twenty-first international conference on
  Machine learning}, page 106. ACM, 2004.

\bibitem{BoWscene}
Jun Yang, Yu-Gang Jiang, Alexander~G. Hauptmann, and Chong-Wah Ngo.
\newblock Evaluating bag-of-visual-words representations in scene
  classification.
\newblock In {\em Proceedings of the International Workshop on Workshop on
  Multimedia Information Retrieval}, MIR '07, 2007.

\end{thebibliography}


@inproceedings{CabralCDBC11,
  author    = {Ricardo S. Cabral and JoÃ£o P. Costeira and Fernando {De la Torre} and Alexandre Bernardino and Gustavo Carneiro},
  title     = {Time and order estimation of paintings based on visual features and expert priors},
  booktitle = {SPIE Electronic Imaging, Computer Vision and Image Analysis of Art II},
  year              = {2011},
}
@inproceedings{Carneiro11,
 author    = {Gustavo Carneiro},
 title     = {Graph-based methods for the automatic annotation and retrieval
              of art prints},
 booktitle = {ICMR},
 year      = {2011},
}
@article{Carneiro11_2,
author = {Carneiro, Gustavo and Costeira, João P.},
title = {The automatic annotation and retrieval of digital images of prints and tile panels using network link analysis algorithms},
year = {2011},
}

@inproceedings{Carneiro12,
 author    = {Gustavo Carneiro and
              Nuno Pinho da Silva and
              Alessio Del Bue and
              Jo{\~a}o Paulo Costeira},
 title     = {Artistic Image Classification: An Analysis on the PRINTART
              Database},
 booktitle = {ECCV},
 year      = {2012},
}

@article{Jia12,
author = {Li, Jia and Yao, Lei and Hendriks, Ella and Wang, James Z.},
title = {Rhythmic Brushstrokes Distinguish van Gogh from His Contemporaries: Findings via Automated Brushstroke Extraction},
journal = {IEEE Trans. Pattern Anal. Mach. Intell.},
year = {2012},
}

@inproceedings{Yelizaveta06,
author = {Yelizaveta, Marchenko and Tat-Seng, Chua and Ramesh, Jain},
title = {Semi-supervised annotation of brushwork in paintings domain using serial combinations of multiple experts},
booktitle = {Proceedings of the 14th annual ACM international conference on Multimedia},
series = {MULTIMEDIA '06},
year = {2006},
}

@inproceedings{PalermoHE12,
 author    = {Frank Palermo and
              James Hays and
              Alexei A. Efros},
 title     = {Dating Historical Color Images},
 booktitle = {ECCV},
 year      = {2012},
}

@inproceedings{Arora12,
  author    = {Ravneet Singh Arora and
               Ahmed M. Elgammal},
  title     = {Towards automated classification of fine-art painting style:
               A comparative study},
  booktitle = {ICPR},
  year      = {2012},
}

@article{Graham10,
author = {Graham, D., Friedenberg, J., Rockmore, D.},
title = {Mapping the similarity space of paintings: image statistics and visual perception},
journal = {Visual Cognition},
year = {2010},
}


@InProceedings{aleb,
 author =    {Lorenzo Torresani and Martin Szummer and Andrew Fitzgibbon},
 title =     {Efficient Object Category Recognition using Classemes},
 booktitle = {ECCV},
 year =      2010,
}

@book{lois,
    author    = "Lois Fichner-Rathus",
    title     = "Foundations of Art and Design",
    publisher = "Clark Baxter",
    year      = "2008",
}

@Book{derek,
author = {Derek Fell},
title = {Van Gogh's Garden},
publisher = {Simon \& Schuster},
year = {2001},
}

@inproceedings{dubuisson,
  title={A modified Hausdorff distance for object matching},
  author={Dubuisson, M-P and Jain, Anil K},
  booktitle={Pattern Recognition},
  year={1994},
}
@article{libsvm,
 author = {Chang, Chih-Chung and Lin, Chih-Jen},
 title = {{LIBSVM}: A library for support vector machines},
 journal = {ACM Transactions on Intelligent Systems and Technology},
 volume = {2},
 issue = {3},
 year = {2011},
 pages = {27:1--27:27},
}

@article{roweis,
  title={Nonlinear dimensionality reduction by locally linear embedding},
  author={Roweis, Sam T and Saul, Lawrence K},
  journal={Science},
  year={2000},
}
@article{Lombardi,
 author    = {Thomas Edward Lombardi},
 title     = {The classification of style in fine-art painting},
 journal = {ETD Collection for Pace University. Paper AAI3189084.},
 year      = {2005},
}

@inproceedings{sablatnig,
  title={Hierarchical Classification of Paintings Using Face- and Brush Stroke Models},
  author={R. Sablatnig, P. Kammerer, and E. Zolda},
  booktitle={ICPR},
  year={1998},
}

@inproceedings{khan,
  title={Who Painted this Painting?},
  author={Fahad Shahbaz Khan, Joost van de Weijer, Maria Vanrell},
  year={2010},
}
@InProceedings{robert, 
  author    =  "Robert {Sablatnig} and Paul {Kammerer} and Ernestine 
                {Zolda}", 
  title     =  "Structural Analysis of Paintings based on Brush
                Strokes", 
  booktitle =  "Proc. of SPIE
                Scientific Detection of Fakery in Art", 
  year      =  "1998", 
  publisher =  "SPIE", 
}  

@inproceedings{widjaja,
  title={Identifying painters from color profiles of skin patches in painting images},
  author={I. Widjaja, W. Leow, and F. Wu.},
  booktitle={ICIP},
  year={2003},
}

@inproceedings{blei,
  title={Latent Dirichlet Allocation},
  author={D. Blei, A. Ng, and M. Jordan.},
  booktitle={Journalof Machine Learning Research},
  year={2003},
}
@INPROCEEDINGS{fei,
    author = {Li Fei-fei},
    title = {A bayesian hierarchical model for learning natural scene categories},
    booktitle = {In CVPR},
    year = {2005},
}
@inproceedings{sivic,
  title={Discovering Objects and their location in images},
  author={Josef Sivic, Bryan C. Russell, Alexei A. Efros, Andrew Zisserman, William T. Freeman},
  booktitle={ICCV},
  year={2005},
}
@ARTICLE{SVM,
    author = {Christopher J. C. Burges},
    title = {A tutorial on support vector machines for pattern recognition},
    journal = {Data Mining and Knowledge Discovery},
    year = {1998},
    volume = {2},
    pages = {121--167}
}
@inproceedings{Csurka2004,
  title={Visual categorization with bags of keypoints},
  author={G. Csurka, C. Dance, L.X. Fan, J. Willamowski, and C. Bray},
  booktitle={Proc. of ECCV International Workshop on Statistical Learning in Computer Vision},
  year={2004},
}
@inproceedings{csift,
  title={CSIFT: A SIFT descriptor with color invariant characteristics},
  author={A. E. Abdel-Hakim and A. A. Farag},
  booktitle={IEEE Conference on Computer Vision and Pattern Recognition, CVPR},
  year={2006},
}
@inproceedings{osift,
  title={Evaluating Color Descriptors for Object and Scene Recognition},
  author={Koen E. A. van de Sande, Theo Gevers and Cees G. M. Snoek},
  booktitle={IEEE Transactions on Pattern Analysis and Machine Intelligence},
  year={2010},
}
@article{sift,
 author = {Lowe, David G.},
 title = {Distinctive Image Features from Scale-Invariant Keypoints},
 journal = {Int. J. Comput. Vision},
 year = {2004},
} 
@INPROCEEDINGS{brdahujapo09,
  author = {Gungor Polatkan, Sina Jafarpour, Andrei Brasoveanu, Shannon Hughes, Ingrid Daubechies}
  year = "2009",
  title = "Detection of forgery in paintings using supervised learning",
  booktitle = "Image Processing (ICIP), 2009 16th IEEE International Conference on",
  month = "nov.",
  pages = "2921 -2924",
}
@MISC{Ng2001,
  author = {Andrew Y. Ng and Michael I. Jordan},
    title = {On Discriminative vs. Generative classifiers: A comparison of logistic regression and naive Bayes},
    year = {2001},
}
@article{VanDeSande2010,
 author = {van de Sande, Koen; Gevers, Theo and Jan-Snoek, Cees G. M.},
 title = {Evaluating color descriptors for object and scene recognition},
 journal = {IEEE Transactions on Pattern Analysis and Machine Intelligence},
volume = {32},
 number = {9},
 year = {2010},
}
@inproceedings{Csurka2004,
  author={G. Csurka, C. Dance, L.X. Fan, J. Willamowski, and C. Bray},
  title={Visual categorization with bags of keypoints},
  booktitle={Proc. of ECCV International Workshop on Statistical Learning in Computer Vision},
  year={2004},
}
@article{ISOMAP,
  author = {Tenenbaum, J. B. and Silva, V. and Langford, J. C.},
  journal = {Science},
  keywords = {classifier clustering dataanalysis method},
  number = 5500,
  pages = {2319--2323},
  title = {{A Global Geometric Framework for Nonlinear Dimensionality Reduction}},
  volume = 290,
  year = 2000
}
@ARTICLE{GIST,
    author = {Aude Oliva and Antonio Torralba},
    title = {Modeling the Shape of the Scene: A Holistic Representation of the Spatial Envelope},
    journal = {International Journal of Computer Vision},
    year = {2001},
    volume = {42},
    pages = {145--175}
}
@InProceedings{HOG,
  author       = "Navneet Dalal and Bill Triggs",
  title        = "Histograms of Oriented Gradients for Human Detection",
  booktitle    = "International Conference on Computer Vision \& Pattern Recognition",
  volume       = "2",
  pages        = "886-893",
  month        = "June",
  year         = "2005",
}
@article{BoW,
 author = {Sivic, Josef and Zisserman, Andrew},
 title = {Efficient Visual Search of Videos Cast As Text Retrieval},
 journal = {IEEE Trans. Pattern Anal. Mach. Intell.},
 issue_date = {April 2009},
} 
@inproceedings{BoW3D,
 author = {Toldo, Roberto and Castellani, Umberto and Fusiello, Andrea},
 title = {A Bag of Words Approach for 3D Object Categorization},
 booktitle = {Proceedings of the 4th International Conference on Computer Vision/Computer Graphics CollaborationTechniques},
 year = {2009},
} 
@inproceedings{BoWscene,
 author = {Yang, Jun and Jiang, Yu-Gang and Hauptmann, Alexander G. and Ngo, Chong-Wah},
 title = {Evaluating Bag-of-visual-words Representations in Scene Classification},
 booktitle = {Proceedings of the International Workshop on Workshop on Multimedia Information Retrieval},
 series = {MIR '07},
 year = {2007}
} 
@ONLINE{artchive,
author = {Harden, Mark},
title = {The Artchive{@http://artchive.com/cdrom.htm}},
url = {http://artchive.com/cdrom.htm/}
}
@ARTICLE{Belkin02laplacianeigenmaps,
    author = {Mikhail Belkin and Partha Niyogi},
    title = {Laplacian Eigenmaps for Dimensionality Reduction and Data Representation},
    journal = {Neural Computation},
    year = {2002},
    volume = {15},
    pages = {1373--1396}
}
@inproceedings{ShiHD09,
  title = {Robust Harris-Laplace Detector by Scale Multiplication},  
  author = {Shi, Fanhuai and Huang, Xixia and Duan, Ye},
  booktitle = {ISVC (1) Lecture Notes in Computer Science},
  date = {2009-12-07},
}
@book{BorgGroenen2005,
  author = {Borg, I. and Groenen, P.J.F.},
  publisher = {Springer},
  title = {{Modern Multidimensional Scaling: Theory and Applications}},
  year = 2005
}

@inproceedings{weinberger2004learning,
  title={Learning a kernel matrix for nonlinear dimensionality reduction},
  author={Weinberger, Kilian Q and Sha, Fei and Saul, Lawrence K},
  booktitle={Proceedings of the twenty-first international conference on Machine learning},
  pages={106},
  year={2004},
  organization={ACM}
}
}
\end{document}